%% file: main.tex
\documentclass[10pt,table,xcdraw]{article} 
\usepackage[preprint]{tmlr}

\usepackage{booktabs}
\usepackage{multicol, multirow}
\usepackage{amsmath}
\usepackage{amssymb}
\usepackage{mathtools}
\usepackage{amsthm}
\usepackage{float}
\usepackage{graphicx}
\usepackage{algorithm}
\usepackage{algorithmic}
\input{math_commands}

\usepackage{hyperref}

\let\classAND\AND
\let\AND\relax

\let\AND\classAND

\AtBeginEnvironment{algorithmic}{\let\AND\algoAND}

\title{Free Lunch for Federated Remote Sensing Target Fine-Grained Classification: A Parameter-Efficient Framework}

\author{\name Shengchao Chen \email pavelchen@ieee.org \\
      \addr Guangdong-Hongkong-Macao Greater Bay Area Weather Research Center for Monitoring Warning and Forecasting (Shenzhen Institute of Meteorological Innovation)\\
      Australian Artificial Intelligence Institute, FEIT, University of Technology Sydney
      \AND
      \name Ting Shu$^\dag$ \email shuting@gbamwf.com \\
      \addr Guangdong-Hongkong-Macao Greater Bay Area Weather Research Center for Monitoring Warning and Forecasting \\ (Shenzhen Institute of Meteorological Innovation)
      \AND
      \name Huan Zhao \email huan.zhao@ntu.edu.sg \\
      \addr School of Electrical and Electronic Engineering, Nanyang Technological University
      \AND
      \name Jiahao Wang \email wangjiahaohainan@163.com \\
      \addr School of Information and Communication Engineering, Hainan University
      \AND
      \name Sufen Ren \email aug\underline{ }0815@163.com \\
      \addr School of Information and Communication Engineering, Hainan University
      \AND 
      \name Lina Yang \email lnyang@gxu.edu.cn \\ 
      \addr School of Computer, Electronics and Information, Guangxi University}

\def\month{MM}  
\def\year{YYYY} 
\def\openreview{\url{https://openreview.net/forum?id=XXXX}} 

\begin{document}

\maketitle

\begin{abstract}
Remote Sensing Target Fine-grained Classification (TFGC) is of great significance in both military and civilian fields. Due to location differences, growth in data size, and centralized server storage constraints, these data are usually stored under different databases across regions/countries. However, privacy laws and national security concerns constrain researchers from accessing these sensitive remote sensing images for further analysis. Additionally, low-resource remote sensing devices encounter challenges in terms of communication overhead and efficiency when dealing with the ever-increasing data and model scales. To solve the above challenges, this paper proposes a novel \textbf{P}rivacy-\textbf{R}eserving TFGC \textbf{F}ramework based on Federated \textbf{L}earning, dubbed \textbf{PRFL}. The proposed framework allows each client to learn global and local knowledge to enhance the local representation of private data in environments with extreme statistical heterogeneity (\textit{non. Independent and Identically Distributed, IID}). Thus, it provides highly customized models to clients with differentiated data distributions. Moreover, the framework minimizes communication overhead and improves efficiency while ensuring satisfactory performance, thereby enhancing robustness and practical applicability under resource-scarce conditions. We demonstrate the effectiveness of the proposed \textbf{PRFL} on the classical TFGC task by leveraging four public datasets.
\end{abstract}

\section{Introduction}
Remote sensing involves detecting and monitoring the characteristics of an area by measuring its radiation from afar, typically through satellites or aircraft~\citep{di2023remote}. This techniques is crucial for both military and civilian uses, as it allows for accurate analysis or images from a distance~\citep{yang2023henc,yi2022mha,yi2023efm}. Target fine-grained classification (TFGC) is a key task in remote sensing image analysis, focusing on identifying subtle differences among similar object categories. Unlike broad category classification, TFGC works to differentiate closely related subcategories, such as various types of ships~\citep{guo2023fine} or aircraft~\citep{zhao2023classification}, which is challenging due to the similarity of their interclass samples and the diversity of features in the intraclass samples.

Existing related works based on advanced Deep Learning (DL) techniques mainly focus on improving the classification performance via introducing well-designed modules and learning strategies to optimize feature extraction capability~\citep{yi2022mha,nurhasanah2023fine,nie2022adap,xiong2022explainable,yi2023efm}. However, these works empirically default to the idea that all data are stored in an idealized central server while ignoring the potential obstacles in real-world applications: privacy and resources. Privacy concerns, along with national and regional security issues, mean that remote sensing data are often considered highly sensitive. This sensitivity can make it challenging to share and work with this data across borders~\citep{zhu2023privacy,zhang2023federated,buyuktacs2023learning}. Regarding resources, edge devices in various locations typically capture remote sensing data. These devices prioritize local storage, but the growing volume of data strains the storage capacity of devices with limited resources\cite{lei2023ultralightweight,buyuktacs2023learning}. Moreover, transferring this data to a central server within a country or region presents additional communication challenges~\citep{chen2023prompt}. These constraints make it difficult for organizations to access a full range of remote sensing data. Consequently, this limits their ability to conduct effective analysis and obtain reliable insights from decision-making systems.

Federated Learning (FL)~\citep{mcmahan2017communication} is a promising learning paradigm that enables multiple clients to train a Machine Learning (ML) model without revealing any private data, such as images and their metadata. It is increasingly popular in fields like healthcare~\citep{dasaradharami2023comprehensive,li2023review}, personalised recommendations~\citep{imran2023refrs,zhang2023comprehensive,zhang2023lightfr}, weather analysis~\citep{chen2023prompt,chen2023spatial}, and remote sensing data analysis~\citep{zhai2023fedleo,zhang2023federated}. Vanilla FL focuses on developing a standard model by frequently updating and sharing the local model parameters among all participants~\citep{mcmahan2017communication}. However, most existing FL methods encounter challenges stemming from the presence of  statistical heterogeneity, characterized non-independent and identically distributed (non-IID) data. This circumstance may markedly hamper the overall performance of the system. This challenge is intensified in federated remote sensing analysis due to the subtle differences between categories and the often skewed distribution of data among those categories. These issues make it difficult to train a global model that performs well across all devices. Finding ways to improve performance amid such pronounced statistical heterogeneity remains an unresolved issue.

Shallow neural networks struggle to capture the intricate representations in remote sensing images. This difficulty arises because image clarity can be compromised by various factors, including noise and irrelevant objects. Researchers have proposed using bigger and more complex network architectures to enhance performance, despite the higher computational costs involved~\citep{xu2023dcsau,yoon2023estimation}. Nonetheless, these complex models create significant challenges within a FL system. Both the server and the clients must exchange model updates frequently during training. Using deeper or wider networks means sending more parameters back and forth, which increases communication demands. Such high communication overhead is particularly problematic for remote devices that often have limited resources.

To address the above issues, this paper proposes a novel \textbf{P}rivacy-P\textbf{R}eserving TFGC \textbf{F}ramework based on Federated \textbf{L}earning, dubbed \textbf{PRFL}. The proposed framework targets cross-regional or international remote sensing TFGC tasks with distributed, heterogeneous data. It enables databases to collaboratively train models personalized to their own data, using knowledge from other devices, without sharing raw data. Within \textbf{PRFL}, we implement a Synchronized Bidirectional Knowledge Distillation (\textbf{SynKD}) strategy, which provides each client with a custom local model that learns from both global and local knowledge without compromising privacy. Additionally, we propose a parameters decomposition method that compresses the full local inference model into low-rank matrices. By applying information entropy-based constraints, we filter out unnecessary parameters. This significantly cuts down communication costs between the server and clients while preserving performance. Consequently, the system trainsmits fewer parameters, making it well-suited for resource-limted remote sensing.

We quantitatively evaluate the performance of the proposed \textbf{PRFL} and typical FL algorithms based on four publicly available remote sensing target fine-grained classification datasets. The main contributions of this work are summarized in four-fold:
\begin{itemize}
    \item We propose a privacy-preserving framework for distributed remote sensing target fine-grained classification tasks. This is the first solution to this problem and addresses the weak performance caused by the notorious heterogeneity while ensuring low communication costs.
    
    \item We propose a simple yet effective knowledge distillation mechanism in \textbf{PRFL}. This mechanism encourages the student model on each client to learn both global and local knowledge while ensuring privacy, thereby tailoring a highly personalized model for each client solving a specific data distribution.
    
    \item We propose dynamic parameter decomposition to reduce communication overhead and improve efficiency by significantly decreasing transmission parameters while maintaining excellent performance.
    
    \item  Extensive experiments on four publicly available real-world TFGC datasets demonstrate that \textbf{PRFL} outperforms state-of-the-art (SOTA) FL algorithms and provides en efficient distributed learning strategy for low-resource scenarios.
\end{itemize}

\section{Related Work}
\subsection{Remote Sensing Target Fine-Grained Classification}
DL have significantly reduced costs and enhanced performance across various fields~\citep{chen2023interpretable,chen2023tempee,chen2022dynamic}, including remote sensing data analysis, where DL excels in feature extraction from raw fragmented data without considering complex physical constrains~\citep{yi2022mha,nurhasanah2023fine,nie2022adap,xiong2022explainable,yi2023efm}. However, remote sensing target fine-grained classification presents a unique challenge: distinguishing subcategories within a base class, complicated by the high similarity among classes and variability within them.

Existing works have utilized deep neural networks to extract features from large-scale data to boost classification accuracy~\citep{nie2022adap,xiong2022explainable,liang2020fgatr,fu2019multicam}, but obtaining distinct features from scattered and imbalanced datasets remains a challenge. Recent efforts have honed in on precise feature extraction in images, using techniques ranging from image pyramid networks with rotated convolutions~\citep{shamsolmoali2021rotation} to methods that blend low-level and high-level semantics~\citep{zhang2021laplacian}, and attention-augmented feature representations~\citep{nie2022adap,liang2020fgatr,zhang2022transformer}.

Despite these advances, current models largely ignore privacy concerns associated with sharing remote sensing data, which may include sensitive information. Our study shifts focus from centralized deep representation extraction to a method that prioritizes privacy without compromising on classification accuracy.

\subsection{Privacy-Preserving Federated Learning}
Federated Learning (FL)~\citep{mcmahan2017communication} is an ML paradigm that allows multiple users to collaboratively train a model without exposing local data~\citep{chen2023collaborative}, and is used in areas such as recommendation systems~\citep{neumann2023privacy,zhang2023lightfr}, climate change~\citep{chen2023prompt,chen2023spatial}, and healthcare~\citep{rauniyar2023federated,zhang2023scoring} due to its privacy-preserving properties. Statistical heterogeneity across clients, also known as non-independent homogeneous distribution (non. IID) in the most important challenge of FL. To handle this problem, several optimization strategies have been proposed and can be categorized into typical FL and personalized FL (pFL).

Vanilla FL assumes that all clients share an identical model as the server. For instance, FedAvg~\citep{mcmahan2017communication} is the first to make this point by updating models in different clients and aggregating their parameters to the server. To solve the issues of non-IID raised in a real-world application, FedProx~\citep{li2020federated} proposed a local regularization term to optimize each client's local model. SCAFFOLD~\citep{karimireddy2020scaffold} theoretically analyzes that the gradient dissimilarity severely hampers the performance of FedAvg and proposes a new stochastic algorithm to overcome this issue via control varieties.

Unlike typical FL, the pFL allows learning different models across clients so naturally suitable for the non-IID challenge. For instance, pFedMe~\citep{t2020personalized} proposes to utilize Moreau envelops as a regularized local objective, which benefits decomposing the personalized model optimization from global learning. Inspired by Model-Agnostic Meta-Learning, Per-FedAvg~\citep{fallah2020personalized} proposes a decentralized version of Meta-Learning to learn a proper initialization model that can be quickly adapted to clients. The aforementioned studies have achieved good performance in pFL, but they have not considered the feature shift issue except FedBN~\citep{li2021fedbn}, which utilizes local batch normalization to alleviate the feature shift. In addition, knowledge distillation-based personalized methods like FedBE~\citep{chen2021fedbe} and FedKD~\citep{wu2022communication} are adopted to reduce the influence of non. IID via distilling global/local representation in CV/NLP tasks.

\section{Preliminaries}
\subsection{Federated Learning}
\label{subsec:fl}
Under Vanilla FedAvg, a central server coordinates $N$ clients\footnote{In this paper, client, device and remote sensing device are synonymous with the mean distributed remote sensing device.} to collaboratively train a uniform global model. Specifically, in each communication round $t$, the server samples a fraction of all the clients, $C$, to join the training. The server distributes the global model $w$ to these clients. Each selected client received global model on their private dataset $D_k ~ P_k$ ($D_k$ obeys the distribution $P_k$) to obtain $w_k$ through local training process $w_k \leftarrow w - \eta \nabla \ell (w; x_i, y_i), (x_i, y_i) \in D_k$. The $k$-th client uploads the trained local model $w_k$ to the server that aggregates them to update the global model via $w = \sum_{k = 0}^{K-1} \frac{n_k}{n} w_k$. FedAvg aims to minimize the average loss of the global model $w$ on all clients' local datasets:
\begin{equation}
    F(w) \text{:}= \mathop{\argmin}\limits_{w_1, w_2, ..., w_N}  \sum_{k=1}^{N} \frac{n_k}{n} F_k(w_k)
\end{equation}
where $n_k$ is the number of samples stored by the $k$-th client. $n$ is the number of samples held by all clients. $F_k(w_k)$ denotes the local objective function of $k$-th client that can be formulated as $\gL_k(w_k) = \ell_k (w_k; (x_i, y_i))$, where the $\gL$ is loss function. 

\subsection{Problem Formulation for PRFL}
The FL setup of our proposed \textbf{PRFL} can be found in Sec.~\ref{subsec:fl}, with the key difference being that each client has unique remote sensing data and feature distribution, creating a statistically heterogeneous environment. Additionally, the remote sensing devices are low-resource edge devices with limited storage capability and communication efficiency. They struggle with processing and transmitting large datasets or complex network. Therefore, \textbf{PRFL} aims to address two main challenges: (1) \textit{How can we minimize the impact of a statistically heterogeneous environment on Federated Remote Sensing TFGC performance in practice?} (2) \textit{How can we reduce communication overhead and improve efficiency without sacrificing performance?}

\begin{figure*}[tbh]
    \centering
    \includegraphics[width=1\textwidth]{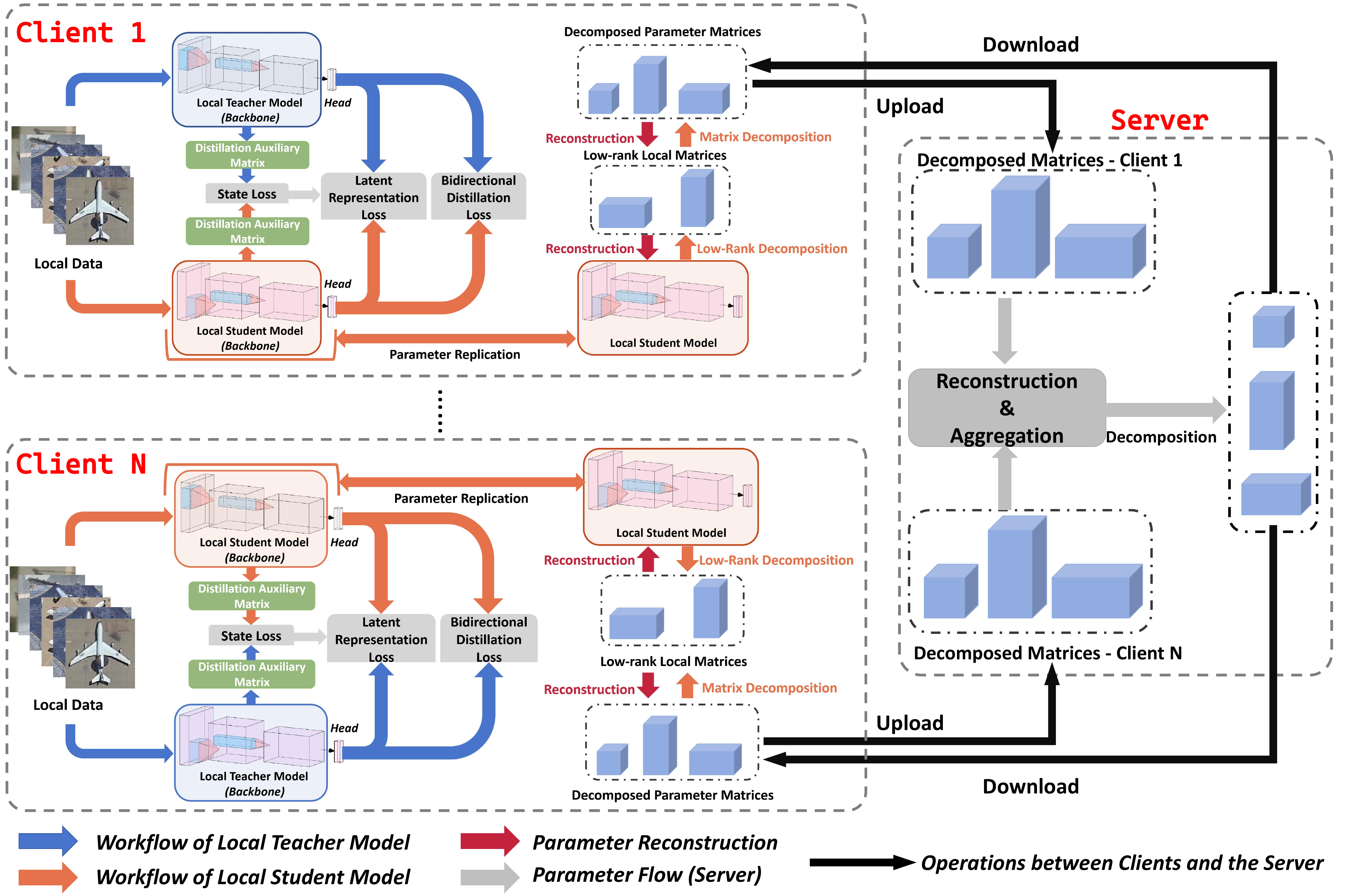}
    \caption{Schematic diagram of the proposed framework.}
    \label{fig:framework}
    \vspace{-10pt}
\end{figure*}

\section{Methodology}
In this section, we elaborate on the details of the proposed framework. This section is divided into three parts, including the architecture of the proposed framework, the details of synchronized bidirectional knowledge distillation and the dynamic parameter decomposition strategy.

\subsection{Framework Architecture}
The architecture of the proposed \textbf{PRFL} system is shown in Fig.~\ref{fig:framework}, comprising $N$ clients, each representing a remote sensing device, and a central server. On each client, there are two local models: a teacher model and a student model. It's important to note that only the student model is active during inference, and it operates solely on local private data to ensure privacy. The server, for security reasons, does not access any client's data.

During each learning iteration, clients calculate updates for their local models and send only the student model's parameters to the server. The server then aggregates these parameters and distributes updated ones back to the clients for the next iteration. This process repeats until the student model converges. This server-client communication approach eliminates the need to share raw data, substantially reducing the amount of private information transmitted compared to traditional centralized training methods. Consequently, this framework offers a degree of privacy protection for clients while training their models.

\subsection{Synchronized Bidirectional Knowledge Distillation}
Knowledge Distillation (KD) allows simpler models to learn from more complex ones, improving their performance by transferring knowledge~\citep{habib2023knowledge}. To tailor models to clients with diverse data distributions and counteract the performance issues caused by non-IID, we've developed a novel KD strategy in \textbf{PRFL} called Synchronized Bidirectional Knowledge Distillation (\textbf{SynKD}). This approach is used locally on each client, as illustrated in Fig.~\ref{fig:framework}, to encourage the local student model accesses the powerful insights both from local and global knowledge.

Unlike conventional KD, our \textbf{SynKD} employs two structurally consistent networks as the teacher and student models. In the client's local update phase, both models learn from the local remote sensing data together. However, only the student model's parameters are upload to the server for collaborative learning, while the teacher model is updated locally. Furthermore, during local updates, the \textbf{SynKD} occurs where both models learn from each other. The teacher model gains from the globally-enhanced student model, enriching its representations. Conversely, the student model benefits from the teacher's local knowledge, improving its personalized representation capabilities. This process helps mitigate the negative effects of non-IID on performance.

Specifically, we introduce two local loss functions to optimize the teacher and student models on each client: the latent representation loss and the bidirectional distillation loss. These functions facilitate the transfer of hidden states and complete model knowledge that's task-oriented. For the latent representation loss, we divide both the teacher and student models into two components: the backbone $\texttt{Backbone}$, which extracts features as hidden states, and the head $\texttt{Head}$, which acts as the classifier. We label the local remote sensing data on client $i$ as $\mX_i$, with predictions from the local teacher and student models noted as $\mY^t_i$ and $\mY^s_i$ respectively. To address the issue of misleading distillation that may arise from inaccurate backbone representations, which can negatively impact performance, we introduce a trainable auxiliary matrix $\mW_{aux}$. This matrix ensures that both the teacher and student models update accurately, enabling a two-way correction between them. The process of latent representation extraction can be described as:
\begin{equation}
\begin{aligned}
    \mH^s_{hs} & = \texttt{Backbone}^{s}(\mX_i), \\
    \mH^t_{hs} & = \texttt{Backbone}^{t}(\mX_i),
\end{aligned}
\end{equation}
where the $\texttt{Backbone}^{s}$ and $\texttt{Backbone}^{t}$ denote the backbone of the teacher and the student model, respectively. Then the process of dual-way correction can be formulated as:
\begin{equation}
    \gL_{cor} = \frac{1}{n} \sum_{i=1}^{n}(\mH^s_{hs} \cdot \mW_{aux} - \mH^t_{hs} \cdot \mW_{aux})^2.
\end{equation}
To construct the above-mentioned latent representation loss, we use the target label to compute the task losses for the teacher model and the student model simultaneously. Task losses for these two models can be formulated based on Cross-Entropy Loss:
\begin{equation}
\begin{aligned}
    \gL^t_{task} = -\sum_{i} \mY^t_i \log(\mY), \\
    \gL^s_{task} = -\sum_{i} \mY^s_i \log(\mY),
\end{aligned}
\label{task:loss}
\end{equation}
where $\gL^t_{task}$ and $\gL^s_{task}$ represent the task loss of the teacher model and student model, repressively, and the $\mY$ is the distribution of ground truth of local label, $\mY^t_i$ and $\mY^s_i$ can be expressed as $\texttt{Head}^s(\mH^s_{hs})$ and $\texttt{Head}^t(\mH^s_{hs})$ according to the above-mentioned \texttt{Backbone-Head} setting. The latent representation loss ($\gL_{lrl}$) for both teacher and student models are formulated below:
\begin{equation}
    \gL^s_{lrl} = \gL^t_{lrl} = \frac{\gL_{cor}}{\gL^t_{task} + \gL^s_{task}}.
\label{eq:latant}
\end{equation}
In this way, the distillation intensity is weak if the prediction of teacher and student models is incorrect (task losses $\gL^t_{task}$ and $\gL^s_{task}$ are large). The distillation is highly effective when the teacher and student models are well trained (task losses $\gL^t_{task}$ and $\gL^s_{task}$ is small), which have the ability to avoid over-fitting and enforcing the extracted latent representation from backbone both teacher and student models toward balance and alignment (have similar knowledge). 

In addition, we further introduce a loss named Bidirectional Distillation Loss that is utilized to distillate the teacher model and student model directly based on their output prediction according to Eq.(\ref{task:loss}), as follows:
\begin{equation}
\begin{aligned}
    \gL^t_{d} = \frac{-\sum_i \mY^t \log (\frac{\mY^s}{\mY^t})}{\gL^t_{task} + \gL^s_{task}}, \\
    \gL^t_{d} = \frac{-\sum_i \mY^s \log (\frac{\mY^t}{\mY^s})}{\gL^t_{task} + \gL^s_{task}},
\end{aligned}
\label{eq:bidiloss}
\end{equation}
where the term of $-\sum_i \mY^t \log (\frac{\mY^s}{\mY^t})$ and $-\sum_i \mY^s \log (\frac{\mY^t}{\mY^s})$ is the Kullback–Leibler divergence (KLD) often appears in conventional knowledge distillation tasks.

Based on the above-proposed loss function, we can now obtain the bidirectional distillation loss for both teacher model (denotes $\gL^t_{bik}$) and student model (denotes $\gL^s_{bik}$) local optimization as follows:
\begin{equation}
\begin{aligned}
    \gL^t_{bik} = \gL^t_{d} + \gL^t_{lrl} + \gL^t_{task},\\
    \gL^s_{bik} = \gL^s_{d} + \gL^s_{lrl} + \gL^s_{task},
\end{aligned}
\end{equation}
Both the teacher and student models are locally updated using the same optimizers on each client. The student model's gradients $\rvg$ on the $i$-th client are calculated as $\rvg^s_i = \frac{\partial \gL^s_{bik}}{\partial \Theta^s}$, where $\Theta^s$ represents the student model's parameters. Similarly, the teacher model's gradients on the same client are obtained using $\rvg^t_i = \frac{\partial \gL^t_{bik}}{\partial \Theta^t}$, with $\Theta^t$ being the teacher model's parameters.

Once both models are updated, each client sends their student model's parameters $\Theta^s$ to the server. The server uses the FedAvg~\citep{mcmahan2017communication} to aggregate these parameters, creating a global model. It then distributes this global model back to the clients. Clients update their student models with the global parameters received from the server. Meanwhile, the local teacher model continues to provide personalized instruction and knowledge transfer. This cycle repeats until both the student and teacher models reach convergence.



\subsection{Dynamic Parameter Decomposition}
The size of the parameters sent from each client often drives the communication overhead in FL. Remote sensing images have complex, fine-grained feature in TFGC tasks, requiring more intricate models to achieve better performance. However, this can be problematic in low-resource environments due to increased communication costs and reduced efficiency. We are exploring the question: \textit{how can we reduce communication costs while still maintaining high performance with minimal or no extra expense?}


To tackle this issue, we introduce a cost-effective method called Dynamic Parameter Decomposition (DPD). DPD compresses the parameters exchanged during server-client communications, aiming to lower communication costs without sacrificing performance, essentially providing a 'free lunch.' We employ a technique known as low-rank parameter matrix decomposition~\citep{hu2021lora} to reduce the burden of transmitting large-scale student model parameters. This step is applied after the local updates of the teacher and student models on the client side, but before the parameters are uploaded, and can be detailed as:

\begin{equation}
\begin{aligned}
    \rvg^s_i &= \rvg^{s,p}_i \cdot \rvg^{s, n}_i, \\
    \text{where } \rvg^s_i \in \sR^{P\times Q}, &\rvg^{s, p}_i \in \sR^{P\times r}, \rvg^{s, n}_i \in \sR^{r\times Q}.
\end{aligned}
\end{equation}
Here\footnote{We use two-dimensional for simplicity.}, the original student model parameters can be split into two low-rank matrices $\rvg^{s, p}_i$ and $\rvg^{s, n}_i$, and $r$ is the rank. In our method, we use a ratio $Q // P$ (if $Q \textgreater P$) or $P // Q$ (if $P \textgreater Q$) as $r$ for simplicity. Before uploading, we decompose the local gradients into smaller matrices using singular value decomposition (SVD). The server then reconstructs the local parameters by multiplying these matrices before aggregating them. The aggregated global parameters are decomposed again and sent to the clients, who then reconstruct them during the model update. Specifically, we apply SVD to the matrices $\rvg^{s,p}_i$ and $\rvg^{s,n}_i$ resulting from the decomposition, which can be expressed as:
\begin{equation}
\begin{aligned}
    \rvg^{s, p}_i \approx \emU^p \Sigma^p \emV^{p}, \\
    \rvg^{s, n}_i \approx \emU^n \Sigma^n \emV^{n}
\end{aligned}
\end{equation}
where $\emU^p \in \sR^{Q \times K}$, $\Sigma^p \in \sR^{K \times K}$, $\emV^p \in \sR^{K \times r}$, $\emU^n \in \sR^{r \times K}$, $\Sigma^n \in \sR^{K \times K}$, $\emV^n \in \sR^{K \times Q}$, and $K$ is the number of retained singular values. In terms of the total number of parameters, if the value of $K$ meets $QK + K^2 + Kr < Pr $ (for $\rvg^{s, p}$), or $rK + K^2 + KQ < Pr $ (for $\rvg^{s, n}$), the uploaded parameter from each client and downloaded parameter from the server can be reduced significantly. In multi-dimensional CNNs utilized in remote sensing image classification, different parameter matrices (e.g., convolution units, linear layers, etc.) are decomposed independently, and the global parameters on the server are decomposed in the same way. We denote the singular values of $\rvg^{s, p}_i$ as $[\sigma_1,\sigma_2,\cdots,\sigma_r]$, $\rvg^{s, n}_i$ as $[\sigma_1,\sigma_2,\cdots,\sigma_Q]$ ordered by their absolute values. To control the approximation error, we propose to combine the Akaike Information Criterion (AIC)~\citep{akaike2011akaike} and the Variance Explained Criterion~\citep{henseler2015new} to determine how many singular values are being transmitted dynamically to maximize the benefits as much as possible while reducing the communication overhead between the server and the client for low-resource scenarios. The method is called AIC constraint-based variance explanation, as detailed below:

\paragraph{Akaike Information Criterion Constraint} AIC can be utilized to balance the goodness of fit and complexity of a model. In the context of FL, model complexity can be understood as the number of parameters that need to be transmitted, while goodness of fit relates to the performance of the model on client data. Therefore, AIC can be represented as follows:
\begin{equation}
    \text{AIC}=2K- 2\ln(L(\theta)),
\end{equation}
where $K$ denotes the number of parameters that need to be transmitted between clients and the server, and $L(\theta)$ is the model likelihood parameterized by local model $\theta$. We use a function to compute the model likelihood and hence singular value selection, which is denoted as:
\begin{equation}
        L(\theta) = P(D|\theta)= \prod_{i=1}^n P(x_i|\theta), 
\end{equation}
where $P(D|\theta)$ is the probability of observing the entire dataset $D$ given the model parameter $\theta$, $P(x_i|\theta)$ denotes the probability of observing a single data point $x_i$ given the model parameter $\theta$. Note that the $D = \{x_1, x_2, ..., x_n\}$ is the training dataset in local clients. In practice, since computing the likelihood function directly may lead to numerical underflow, we usually turn to the logarithmic form of the likelihood, i.e., the log-likelihood function:
\begin{equation}
    \ln(L(\theta)) = \sum_{i=1}^n \ln(P(x_i|\theta)).
\end{equation}
Therefore, AIC can be formulated as:
\begin{equation}
    \text{AIC}=2K- 2\sum_{i=1}^n \ln(P(x_i|\theta)).
\end{equation}
\paragraph{AIC-constraint Variance Explained} Based on AIC, we can rely on the variance explained to accuracy and use AIC to constrain its optimization, further balancing the number of transmitted parameters and model performance. This equation can be formulated as:
\begin{equation}
\begin{aligned}
    &\min_K \frac{\sum_{i=1}^K \sigma_i^2}{\sum_{i=1}^Q \sigma_i^2} \textgreater \alpha, \\
    & s.t. \quad \min_K [K- \sum_{i=1}^n \ln(P(x_i|\theta))].
\end{aligned}
\end{equation}
With the AIC-constrained Variance Explained strategy, clients can transmit parameters that hold as much relevant information as possible when interacting with the server. This approach helps prevent the negative effects of sending unnecessary parameters on both model performance and communication costs. Additionally, the parameters transmitted are not static; they are dynamically selected based on how well they perform on the client's private data, adding flexibility to the method. The algorithm implement of the proposed \textbf{PRFL} is shown in Alg.~\ref{alg1}.

\begin{algorithm}[tb]
   \caption{Algorithm Implement of PRFL}
   \label{alg:example}
    \begin{algorithmic}
   \STATE Setting the local learning rate $\eta$, client number $N$, and local dataset $D_i=\{x_1, x_2, ..., x_n\}$
   \STATE Setting communication rounds $R$, local updating steps $e$
   \STATE Setting hyperparameters $\alpha$,
   \FOR{communication rounds $R=1,2,3...$}
   \STATE \colorbox{green! 40}{\bf Client-side:}
   \FOR{each client $i$ in parallel}
   \STATE Initialize student model $\Theta^s_i$ and teacher model $\Theta^t_i$
   \STATE $\Theta^t_i \leftarrow \textsc{LocalUpdate}(D_i, \eta, \Theta^t_i/ \rvg^t_i, e)$ 
   \STATE $\Theta^s_i \leftarrow \textsc{LocalUpdate}(D_i, \eta, \Theta^s_i, e)$ 
   \STATE $\rvg^{s,p}_i, \rvg^{s,n}_i  \leftarrow \textsc{Decom}(\rvg^s_i, e)$
   \STATE $\emU^p \Sigma^p \emV^{p} \leftarrow \textsc{DPD}(\rvg^{s,p}_i, e)$
   \STATE $\emU^n \Sigma^n \emV^{n} \leftarrow \textsc{DPD}(\rvg^{s,n}_i, e)$
   \STATE Clients upload $\emU^p_i \Sigma^p_i \emV^{p}_i$ and $\emU^n_i \Sigma^n_i \emV^{n}_i$ to the server
   \ENDFOR
   \STATE \colorbox{red! 40}{\bf Server-side:}
   \STATE Server receives $\emU^p \Sigma^p \emV^{p}$ and $\emU^n \Sigma^n \emV^n$
   \STATE Server reconstructs $\rvg^t_i$
   \STATE Server aggregates $\rvg \leftarrow \sum_{i}^{N} \frac{n_i}{n}\rvg^t_i$
   \STATE $\rvg \leftarrow \emU \Sigma \emV$
   \STATE Server distribute $\rvg \leftarrow \emU \Sigma \emV$ to each client
   \ENDFOR
   \STATE \colorbox{black! 40}{\bf LocalUpdate$(D, \eta, \Theta, e)$:}
   \FOR{local updating steps $e=1,2,3...$}
   \STATE Local model $\Theta^s$ and $\Theta^t$ trained by private data $D$
   \STATE Compute task losses according to Eq.~\ref{task:loss}
   \STATE Compute latent representation losses according to Eq.~\ref{eq:latant}
   \STATE Compute distillation losses according to Eq.~\ref{eq:bidiloss}
   \ENDFOR
\end{algorithmic}
\label{alg1}
\end{algorithm}

\section{Experiments and Results}
In this section, we describe the dataset and the experimental environment setup. We then present the results conducted in various Non-IID settings. These include standard experiment reports, ablation studies on key components, assessments of parameter impacts, and experiments on differential privacy.

\subsection{Dataset}
Four publicly available remote sensing image datasets for target fine-grained classification, including MTARSI, MSTAR, FGSCR-42, and Aircraft-16, are used to evaluate the performance of our proposed framework. Their detail information is shown in Table~\ref{tab:dataset_stat} and as below:

\textbf{MTARSI}\footnote{\url{https://www.kaggle.com/datasets/aqibriaz/mtarsidataset}} is a dataset that contains 9598 images of 20 aircraft categories from remote seining images, with complex background, different spatial resolution, and pose variations.

\textbf{MSTAR}\footnote{\url{https://www.kaggle.com/datasets/atreyamajumdar/mstar-dataset-8-classes}} is a dataset that consists of eight fine-grained categories of military vehicles in the X-band and HH polarization SAR images with the spatial resolution of 0.3m, note that we only use SAR images with eight classes in our experiments.

\textbf{FGSCR-42}\footnote{\url{https://github.com/DYH666/FGSCR-42}}~\citep{di2021public} is a dataset for fine-grained ship classification from optical remote sensing images of 42 ship categories (e.g., Kitty-Hawk class aircraft carrier, Arleigh-Burke class destroyer, and mega-yacht etc.). It contains 9320 images whose sizes range from about $50 \times 50$ to about $1500 \times 1500$ pixels with various aspect ratios.

\textbf{Aircraft-16}\footnote{\url{https://github.com/JACYI/Dataset-for-Remote-Sensing}}~\citep{yi2022mha} contains 8,488 images from 16 different aircraft classes. The challenge with the dataset is the similarity of the airline between aircraft classes.

\begin{table}[tbh]
    \centering
    \resizebox{.65\textwidth}{!}{
    \begin{tabular}{cccc}
    \toprule
        Dataset & \#Classes & \#Images & \#Devices (Clients)\\
    \midrule
        MTARSI &  20& 9,598 & 20\\
        MSTAR &  8 & 9466 & 20\\
        FGSCR-42 &  42& 9,320 & 20\\
        Aircraft-16 &  16& 8,488 & 20\\
    \bottomrule
    \end{tabular}}
    \caption{Statistics of remote sensing image fine-grained classification datasets, note that the \# Devices (Clients) is a simulated parameter in our FL setting.}
    \label{tab:dataset_stat}
\end{table}
\vspace{-10pt}

\subsection{Dataset Pre-Processing}
\begin{figure}[tbh]
    \centering
    \includegraphics[width=.8\textwidth]{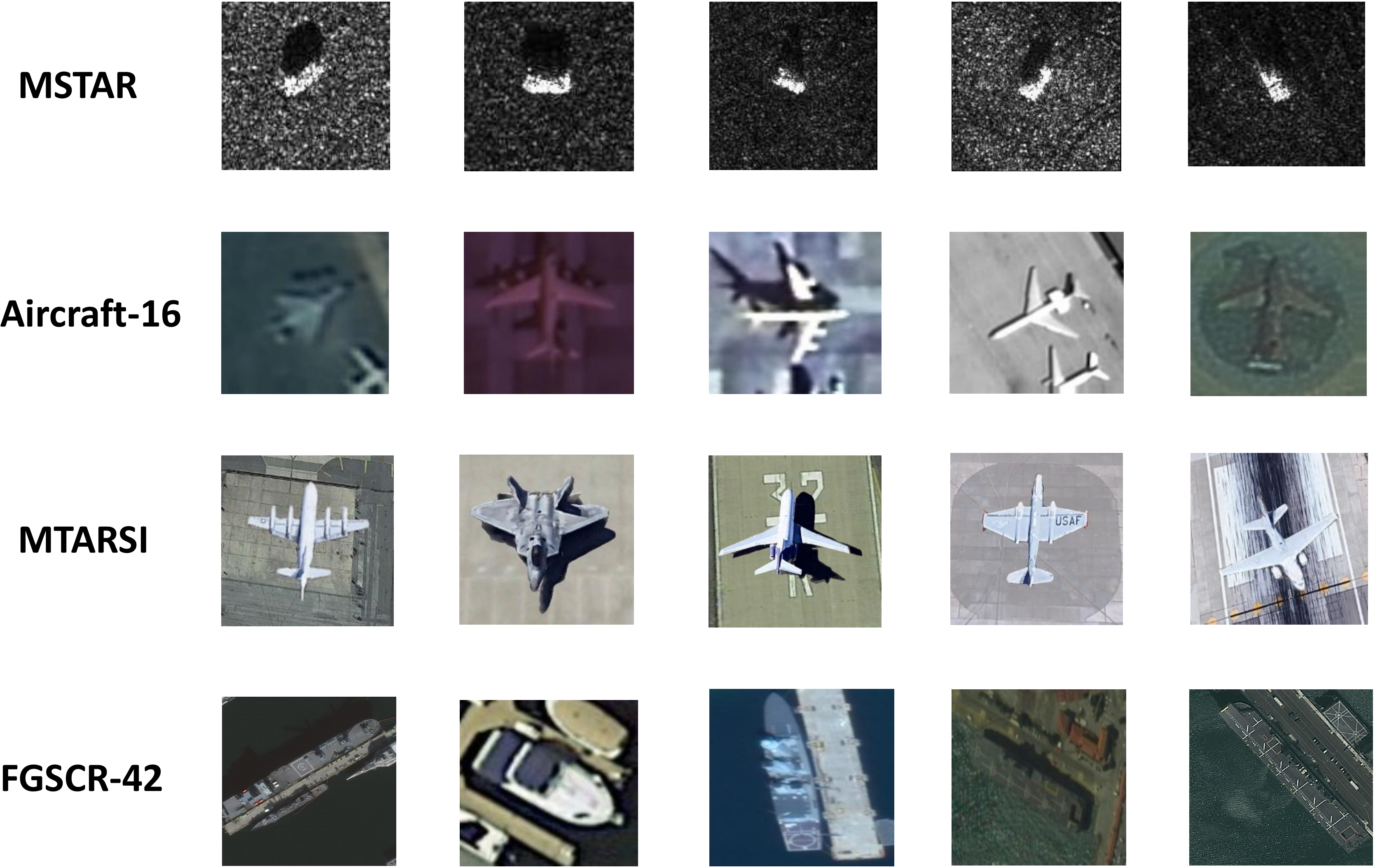}
    \caption{Partial visualization of the benchmark dataset used in this study. From top to bottom are MSTAR, Aircraft-16, MTARSI, and FGSCR-42. It is clear that the distributions of the different labeled samples in these datasets are extremely similar.}
    \label{fig:dataset}
\end{figure}

For the datasets mentioned, we applied uniform data preprocessing. We first resized all images to a standard resolution of $299 \times 299$, as shown in Fig.~\ref{fig:dataset}. Then, we distributed these images across 20 clients to mimic the FL environment. On each client, we divided the data into training, testing, and validation sets in the ratios of 80\%, 10\%, and 10\%, respectively.

To verify our proposed framework's effectiveness in various non-IID scenarios, we employed two common data distribution approaches as described in Ref.~\citep{wu2023bold}:

\textbf{Pathological Non-IID Setting.} This setting, which is frequently used in FL, assigns two classes of data to each client at random, creating a different sample distribution for each. Consequently, each client has a simple binary classification task to perform locally.

\textbf{Dirichlet Non-IID Setting.} Another typical FL scenario is simulated by drawing each client's data from a Dirichlet distribution, expressed as $\rvq \sim \texttt{Dirichlet}(\lambda \rvp)$. Here, $\rvp$ is the prior class distribution, and $\lambda$ is a factor that adjusts the degree of non-IID. A larger $\lambda$ indicates substantial class imbalance within each client, leading to more challenging local tasks, such as a greater variety of classes and fewer samples per class. As $\lambda$ increases, the differences in data distribution among clients decrease, but the variety within client data distribution grows. This setting effectively tests the robustness of methods in complex Non-IID environments. We experimented with two different Dirichlet parameters, $\{0.02, 0.1\}$, to explore varying non-IID scenarios.

\subsection{Baselines}
We have selected the well-known SOTA FL algorithms, including a personalized algorithm based on parameter localized (\textsc{FedBN}), based on local regularization (\textsc{PerFedAvg}), and based on knowledge distillation (\textsc{FedBE}) as the baseline, to demonstrate the superiority of our proposed framework in the federated remote sensing images fine-grained classification task. Their detailed information is as follows:

\textbf{\textsc{Local.}} Without any distributed learning environment all of the remote sensing images are trained in a centralized manner.

\textbf{\textsc{FedAvg.}}~\citep{mcmahan2017communication} Aggregating local models to obtain a global model via classic average strategy while preserving the privacy of each individual's data.

\textbf{\textsc{FedBN.}}~\citep{li2021fedbn} A personalized FL approach that personalized batch-norm layers in local models and shares remaining parameters globally to achieve highly customized models for each client.

\textbf{\textsc{PerFedAvg.}}~\citep{t2020personalized} A personalized FL approach that adapts the global model to each user’s local data distribution while taking into account the similarity between users to improve model generalization.

\textbf{\textsc{FedBE.}}~\citep{chen2021fedbe} A personalized FL approach using Bayesian model integration for robustness in aggregating user predictions and summarizing integrated predictions into global models with the help of knowledge distillation.

\textbf{\textsc{PRFL.}} The proposed framework introduces Synchronized Bidirectional Knowledge Distillation and a cost-effective parameters decomposition to achieve a powerful global and highly customized local model for each client under privacy-persevering conditions.

\begin{figure*}[tbh]
    \centering
    \includegraphics[width=1\textwidth]{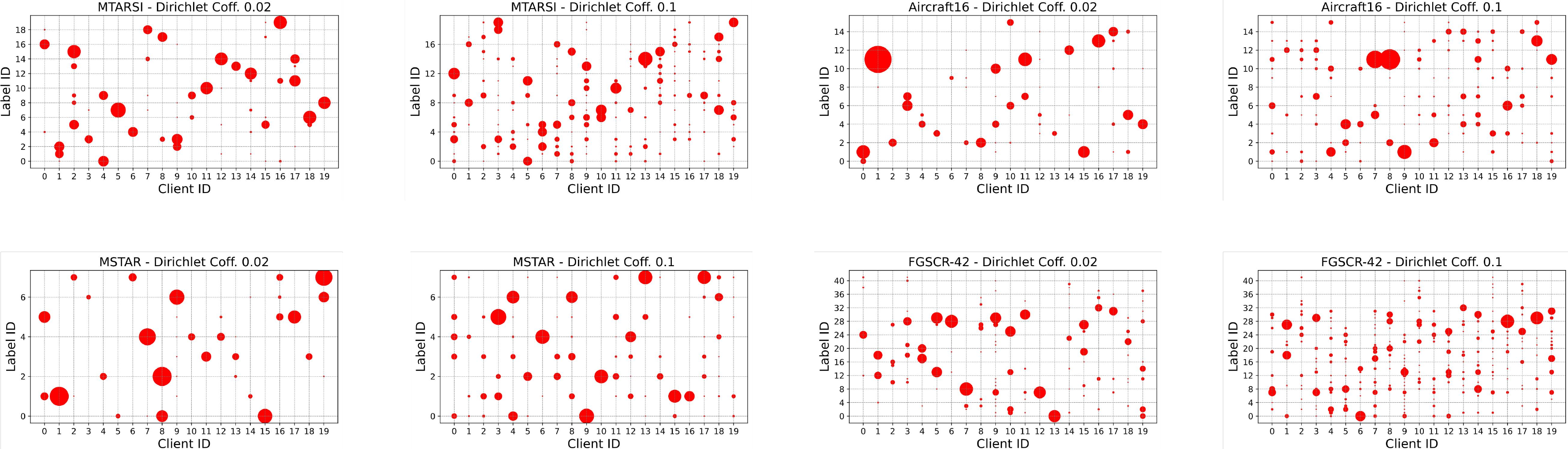}
    \caption{Visualization of four benchmark datasets. From top to bottom are \textsc{MTARSI}, \textsc{Aircraft-16}, \textsc{MSTAR}, and \textsc{FGSCR-42}, where a larger red circle means that the client has more such samples, and the opposite means that there are fewer such samples on the client.}
    \label{fig:dataset_split}
\end{figure*}

\begin{table}[htbp]
  \centering
  \resizebox{1\textwidth}{!}{
    \begin{tabular}{c|ccc|ccc|ccc|ccc}
    \toprule
    \multicolumn{13}{c}{\cellcolor{green! 15} \textsc{Pathological Non-IID Setting}} \\
    \midrule
    \textsc{Dataset} & \multicolumn{3}{c|}{MTARSI} & \multicolumn{3}{c|}{MSTAR} & \multicolumn{3}{c|}{FGSCR-42} & \multicolumn{3}{c}{Aircraft-16} \\
    \midrule
    Method/Ratio & 10\% & 30\% & 50\% & 10\% & 30\% & 50\% & 10\% & 30\% & 50\% & 10\% & 30\% & 50\% \\
    \midrule
    \multirow{2}[0]{*}{\textsc{FedAvg}} & 59.60  & 69.55  & 71.65  & 49.60  & 52.55  & 55.40  & 67.20  & 77.15  & 81.25  & 56.10  & 65.50  & \bf 91.30  \\
       & \small $\pm0.42$ & \small $\pm0.56$ & \small $\pm0.29$ & \small $\pm0.41$ & \small $\pm0.59$ & \small $\pm0.42$ & \small $\pm1.11$ & \small $\pm0.09$ & \small $\pm0.27$ & \small $\pm0.63$ & \small $\pm0.24$ & \small $\pm0.49$ \\
    \multirow{2}[0]{*}{\textsc{Local}} & 39.95  & 76.20  & 85.40  & 57.75  & 87.35  & \underline{95.55}  & 39.80  & 73.85  & 80.30  & 39.70  & 76.55  & 81.75  \\
       & \small $\pm1.23$ & \small $\pm0.20$ & \small $\pm0.18$ & \small $\pm0.42$ & \small $\pm0.92$ & \small $\pm0.42$ & \small $\pm0.92$ & \small $\pm1.45$ & \small $\pm1.12$ & \small $\pm0.52$ & \small $\pm0.65$ & \small $\pm0.69$ \\
    \multirow{2}[0]{*}{\textsc{FedBN}} & 75.55  & \bf 94.17  & \underline{94.42}  & 68.20  & 84.10  & 87.75  & \bf 80.95  & \underline{81.30}  & \underline{82.78}  & 63.95  & \underline{83.30}  & 86.98  \\
       & \small $\pm0.44$ & \small $\pm0.19$ & \small $\pm1.25$ & \small $\pm1.37$ & \small $\pm1.24$ & \small $\pm0.47$ & \small $\pm1.72$ & \small $\pm0.46$ & \small $\pm0.28$ & \small $\pm0.48$ & \small $\pm0.47$ & \small $\pm0.21$ \\
    \multirow{2}[0]{*}{\textsc{PerFedAvg}} & 80.35  & 86.40  & 90.60  & 88.35  & \underline{92.25}  & 91.70  & 77.98  &  79.21  &  81.21  & 71.05  & 76.60  & 76.35  \\
       & \small $\pm0.16$ &  \small $\pm0.47$ & \small $\pm0.96$ & \small $\pm1.07$ & \small $\pm0.55$ & \small $\pm0.38$ & \small $\pm0.92$ & \small $\pm1.70$ & \small $\pm1.42$ & \small $\pm0.53$ & \small $\pm0.37$ & \small $\pm0.88$ \\
    \multirow{2}[0]{*}{\textsc{FedBE}} & \underline{84.01}  & 84.18  & 85.26  & \underline{86.14}  & 86.99  & 88.04  & 75.44  & 77.25   &  80.19  & \underline{78.42}  & 81.55  & 81.95  \\
       & \small $\pm1.71$ & \small $\pm0.32$ & \small $\pm0.58$ & \small $\pm1.25$ & \small $\pm0.27$ & \small $\pm0.43$ & \small $\pm1.23$ &  \small $\pm1.24$  &  \small $\pm1.49$  & \small $\pm0.70$ & \small $\pm2.01$ & \small $\pm0.36$ \\
       \midrule
    \multirow{2}[0]{*}{\textbf{\textsc{PRFL (Ours)}}} & \bf 89.85  & \underline{93.72}  & \bf  94.47  & \bf 96.68  & \bf 97.82  & \bf 97.11  & \underline{80.20}  & \bf 84.35  & \bf 85.19  & \bf 84.24  & \bf 86.55  & \underline{89.27} \\
       & \small $\pm0.37$ & \small $\pm0.90$ & \small $\pm1.00$ & \small $\pm1.02$ & \small $\pm0.21$ & \small $\pm0.42$ & \small $\pm0.41$ & \small $\pm0.29$ & \small $\pm0.43$ & \small $\pm0.29$ & \small $\pm0.47$ & \small $\pm0.34$ \\
    \bottomrule
    \end{tabular}}
    \caption{Experiment results of the proposed PRFL and baseline under Pathological Non-IID setting for four fine-grained classification datasets with client participant ratio $\rvr \in \{10\%, 30\%, 50\%\}$, where the \textbf{Bold} indicates the optimal results, and \underline{Underline} represent the suboptimal result.}
  \label{tab:path_exp_res}
\end{table}

To ensure a fair comparison, we keep all baseline parameters the same as those in the original publication.
For local model on each client, we select \textbf{ResNet18}~\citep{he2016deep} pretrained with ImageNet~\citep{deng2009imagenet} for better initialization. During FL training, we set the number of global communication rounds to 100 and local update epochs to 5. We use SGD~\citep{ruder2016overview} as the optimizer with a local learning rate of $5e^{-3}$ for local teacher and student models. The framework's performance is evaluated using accuracy (\textbf{ACC}).

Additionally, we introduce a parameter $\rvr$ to regulate the number of clients that participate in each communication round. This allows us to assess our framework's effectiveness and advantage in low-resource situations, where server communication capacity limits may only permit a small number of clients to be active each round. For instance, with $\rvr = 0.1$, only 2 clients ($\rvr \times 20$) are randomly chosen to participate in the training per round. We conducted all experiments five times in various randomized settings and reported the average and variability of the results.

\begin{table}[tbh]
  \centering
  \resizebox{1\textwidth}{!}{
    \begin{tabular}{c|ccc|ccc|ccc|ccc}
    \toprule
    \multicolumn{13}{c}{\cellcolor{green! 15} \textsc{Dirichlet Non-IID Setting ($\lambda=0.1$)}} \\
    \midrule
    \textsc{Dataset} & \multicolumn{3}{c|}{MTARSI} & \multicolumn{3}{c|}{MSTAR} & \multicolumn{3}{c|}{FGSCR-42} & \multicolumn{3}{c}{Aircraft-16} \\
    \midrule
    Method/Ratio & 10\% & 30\% & 50\% & 10\% & 30\% & 50\% & 10\% & 30\% & 50\% & 10\% & 30\% & 50\% \\
    \midrule
    \multirow{2}[0]{*}{\textsc{FedAvg}} & 57.55  & 67.48  & 69.60  & 47.56  & 50.53  & 53.37  & 65.15  & 76.10  & 79.22  & 54.08  & 63.43  & \underline{90.23}  \\
       & \small $\pm1.21$ & \small $\pm1.67$ & \small $\pm1.80$ & \small $\pm0.71$ & \small $\pm1.45$ & \small $\pm2.10$ & \small $\pm2.11$ & \small $\pm1.01$ & \small $\pm0.24$ & \small $\pm1.83$ & \small $\pm0.55$ & \small $\pm3.12$ \\
    \multirow{2}[0]{*}{\textsc{Local}} & 37.92  & 74.18  & 83.37  & 55.69  & 85.30  & \underline{94.48}  & 37.72  & 71.81  & 78.25  & 37.65  & 74.51  & 79.72  \\
       & \small $\pm0.93$ & \small $\pm1.88$ & \small $\pm0.76$ & \small $\pm1.11$ & \small $\pm1.59$ & \small $\pm1.84$ & \small $\pm0.38$ & \small $\pm1.40$ & \small $\pm1.22$ & \small $\pm0.67$ & \small $\pm1.41$ & \small $\pm2.11$ \\
    \multirow{2}[0]{*}{\textsc{FedBN}} & 73.51  & \underline{92.52}  & \underline{93.11}  & 66.13  & 82.06  & 85.68  & \underline{81.93}  & \underline{88.28}  & \underline{94.73}  & 61.92  & \underline{81.27}  & 84.93  \\
       & \small $\pm2.04$ & \small $\pm1.23$ & \small $\pm2.41$ & \small $\pm0.06$ & \small $\pm1.11$ & \small $\pm0.57$ & \small $\pm1.12$ & \small $\pm2.76$ & \small $\pm0.98$ & \small $\pm1.81$ & \small $\pm2.27$ & \small $\pm1.54$ \\
    \multirow{2}[0]{*}{\textsc{PerFedAvg}} & 78.30  & 84.37  & 88.55  & 86.31  & \underline{90.23}  & 89.68  & 75.95  &  76.21  &  79.77  & 69.01  & 74.55  & 74.32  \\
       & \small $\pm0.97$ & \small $\pm1.33$ & \small $\pm0.42$ & \small $\pm0.54$ & \small $\pm0.69$ & \small $\pm0.93$ & \small $\pm1.98$ & \small $\pm0.92$   &  \small $\pm1.27$  & \small $\pm1.29$ & \small $\pm2.56$ & \small $\pm0.18$ \\
    \multirow{2}[0]{*}{\textsc{FedBE}} & \underline{85.10}  & 85.88  & 86.94  & \underline{88.94}  & 89.20  & 89.27  & 78.45  & 80.58   &  80.79  & \underline{79.62}  & 80.38  & 80.61  \\
       & \small $\pm1.22$ & \small $\pm1.92$ & \small $\pm1.03$ & \small $\pm0.71$ & \small $\pm1.07$ & \small $\pm0.93$ & \small $\pm1.10$ &  \small $\pm1.44$  &  \small $\pm1.71$  & \small $\pm0.75$ & \small $\pm1.01$ & \small $\pm1.22$ \\
       \midrule
    \multirow{2}[0]{*}{\textbf{\textsc{PRFL (Ours)}}} & \bf 90.35  & \bf 94.22  & \bf 94.97  & \bf 97.18  & \bf 98.82  & \bf 99.12  & \bf 84.49  & \bf 93.35  & \bf 89.05  & \bf 87.04  & \bf 89.48  & \bf 90.33  \\
       & \small $\pm0.79$ & \small $\pm1.73$ & \small $\pm1.09$ & \small $\pm0.15$ & \small $\pm0.84$ & \small $\pm0.42$ & \small $\pm0.91$ & \small $\pm1.12$ & \small $\pm0.23$ & \small $\pm0.87$ & \small $\pm1.38$ & \small $\pm1.92$ \\
    \bottomrule
    \end{tabular}}
      \caption{Experimental results of the proposed PRFL and baseline under Dirichlet Non-IID setting for four fine-grained image classification datasets with $\lambda=0.1$ and client participant ratio $\rvr \in \{10\%, 30\%, 50\%\}$, where \textbf{Bold} are the optimal results, and \underline{underline} represent suboptimal result.}
  \label{tab:dir_exp_dot1}
\end{table}

\begin{table}[htbp]
  \centering
  \resizebox{1\textwidth}{!}{
    \begin{tabular}{c|ccc|ccc|ccc|ccc}
    \toprule
    \multicolumn{13}{c}{\cellcolor{green! 15} \textsc{Dirichlet Non-IID Setting ($\lambda=0.02$)}} \\
    \midrule
    \textsc{Dataset} & \multicolumn{3}{c|}{MTARSI} & \multicolumn{3}{c|}{MSTAR} & \multicolumn{3}{c|}{FGSCR-42} & \multicolumn{3}{c}{Aircraft-16} \\
    \midrule
    Method/Ratio & 10\% & 30\% & 50\% & 10\% & 30\% & 50\% & 10\% & 30\% & 50\% & 10\% & 30\% & 50\% \\
    \midrule
    \multirow{2}[0]{*}{\textsc{FedAvg}} & 30.05  & 37.86  & 39.03  & 37.14  & 48.42  & 48.25  & 41.39  & 46.06  & 50.56  & 38.12  & 42.25  & 42.02  \\
       & \small $\pm 1.13$ &  \small $\pm 0.42$ & \small $\pm 1.26$ & $\pm 2.77$ & \small $\pm 0.78$ & \small $\pm 0.93$ & \small $\pm 0.65$ & \small $\pm1.98$ & \small $\pm1.45$ & \small $\pm2.21$ & \small $\pm1.29$ & \small $\pm2.21$ \\
    \multirow{2}[0]{*}{\textsc{Local}} & 44.51  & 84.62  & \underline{89.32}  & 63.62  & 86.15  & \underline{96.61}  & 41.55  & 77.67  & 87.19  & 50.89  & 82.06  & \underline{94.79}  \\
       & \small $\pm2.76$ & \small $\pm2.14$ & \small $\pm1.37$ & \small $\pm1.61$ & \small $\pm1.23$ & \small $\pm0.84$ & \small $\pm1.46$ & \small $\pm2.05$ & \small $\pm0.64$ & \small $\pm1.93$ & \small $\pm0.39$ & \small $\pm1.05$ \\
    \multirow{2}[0]{*}{\textsc{FedBN}} & 38.36  & 56.57  & 61.80  & 52.72  & 58.82  & 61.31  & 53.07  & 66.34  & 73.41  & 39.15  & 56.81  & 73.65  \\
       & \small $\pm0.45$ & \small $\pm1.87$ & \small $\pm0.09$ & \small $\pm2.64$ & \small $\pm0.81$ & \small $\pm2.32$ & \small $\pm0.87$ & \small $\pm2.51$ & \small $\pm1.14$ & \small $\pm2.26$ & \small $\pm1.47$ & \small $\pm1.11$ \\
    \multirow{2}[0]{*}{\textsc{PerFedAvg}} & 60.64  & 79.84  & 84.70  & 75.37  & 86.44  & 91.28  & 72.18  & 86.44  & 83.25  &  71.21  &  82.48  & 84.06 \\
       & \small $\pm0.82$ & \small $\pm1.96$ & \small $\pm1.68$ & \small $\pm0.54$ & \small $\pm1.19$ & \small $\pm0.23$ & \small $\pm0.93$ & \small $\pm0.37$ & \small $\pm2.75$ & \small $\pm0.64$   &  \small $\pm1.10$  & \small $\pm0.51$ \\
    \multirow{2}[0]{*}{\textsc{FedBE}} & \underline{89.10}  &  \underline{89.21}  & 89.17  & \underline{89.91}  & \underline{90.11}  &  90.87 & \underline{88.62}  &  \underline{89.13}  &  \underline{89.12}  & \underline{87.28}  &  \underline{87.32} & 87.60  \\
       & \small $\pm1.46$ & \small $\pm1.21$ & \small $\pm1.90$ & \small $\pm0.88$ & \small $\pm1.17$ & \small $\pm1.13$ & \small $\pm2.10$ &  \small $\pm1.27$  &  \small $\pm1.26$  & \small $\pm0.77$ & \small $\pm1.21$ & \small $\pm1.37$ \\
    \midrule
    \multirow{2}[0]{*}{\textbf{\textsc{PRFL (Ours)}}} & \bf 97.01  & \bf 98.84  & \bf 98.92  & \bf 98.28  & \bf 98.79  & \bf 98.83  & \bf 96.31  & \bf 97.85  & \bf 98.26  & \bf 96.62  & \bf 97.18  & \bf 97.23  \\
       & \small $\pm0.98$ & \small $\pm0.27$ & \small $\pm0.46$ & \small $\pm0.78$ & \small $\pm0.42$ & \small $\pm0.91$ & \small $\pm0.62$ & \small $\pm0.19$ & \small $\pm0.99$ & \small $\pm1.46$ & \small $\pm1.73$ & \small $\pm1.12$ \\
    \bottomrule
    \end{tabular}}
    \caption{Experimental results of the proposed PRFL and baseline under Dirichlet Non-IID setting for four fine-grained image classification datasets with $\lambda=0.02$ and client participant ratio $\rvr \in \{10\%, 30\%, 50\%\}$, where \textbf{Bold} are the optimal results, and \underline{underline} represent suboptimal result.}
  \label{tab:dir_exp_dot2}
\end{table}
\subsection{Experiments}
We conduct experiments on Dirichlet Non-IID and Pathological Non-IID setting for each dataset to demonstrate the superiority and effectiveness of our proposed method, respectively. Detailed analysis and discussion are as below according to the categorization of Non-IID setting.

\paragraph{Results on Pathological Non-IID Setting} The experimental results in the pathological non-IID setting are presented in Table~\ref{tab:path_exp_res}. The findings indicate that: \textbf{(1)} Our proposed PRFL outperforms the baseline methods in most cases across four datasets, demonstrating greater robustness. \textbf{(2)} Notably, in specific instances such as testing different algorithms on the Aircraft-16 dataset, PRFL falls short of the traditional FL algorithm FedAvg in accuracy when 50\% of the clients participate in training. However, with 10\% and 30\% client participation in training, PRFL significantly outshines the competition. The Aircraft-16 dataset, under the pathological non-IID setting, has a limited number of categories (just 16). This can lead to an overlap in sample categories among clients, especially given the high similarity within the dataset's categories. Consequently, the increase in samples from one to two per client in the Aircraft-16 dataset narrows the sample distribution differences, thus reducing the non-IID severity. Although the MTARSI and MSTAR datasets also have a limited number of samples, they don't exhibit the same level of category overlap as seen with Aircraft-16. Further analysis of why PRFL excels will be discussed in the context of the Dirichlet setting experiments.

\paragraph{Results on Dirichlet Non-IID setting} We set the Dirichlet control coefficients to $\lambda=0.02$ and $\lambda=0.1$ to mimic a non-IID environment. The data distribution for each client in four datasets is shown in Fig.~\ref{fig:dataset_split}.

Our PRFL's experimental results, conducted under a Dirichlet non-IID setting with $\lambda=0.1$, are compared across four datasets with varying client participation ratios. Table~\ref{tab:dir_exp_dot1} shows that PRFL consistently outperforms baseline methods. Specifically: \textbf{(1)} On the MTARSI dataset, PRFL reaches 90.35\% accuracy with only 10\% of clients involved in training, a 5.25\% absolute increase over the next best baseline. This gap widens to 8.34\% and 8.03\% improvements at 30\% and 50\% client participation, respectively, when compared to FedBE; \textbf{(2)} PRFL maintains its lead on the FGSCR-42 dataset, achieving 89.05\% accuracy at a 50\% labeling ratio. This is a notable 4.32\% absolute and 5.10\% relative improvement over FedBN's results; \textbf{(3)} PRFL shows stable performance across all datasets, as indicated by lower standard deviations compared to other state-of-the-art methods, suggesting it is more reliable.

Under the more challenging $\lambda=0.02$, which implies greater heterogeneity, PRFL's superiority persists, as shown in Table~\ref{tab:dir_exp_dot2}. Unlike FedAvg, FedBN, and PerFedAvg, which suffer under increased class imbalance, FedBE and PRFL improve due to their distillation mechanisms. These enable clients to learn knowledge globally, combining it with local data for better personalization, thus countering the heterogeneity's adverse effects.

PRFL specifically outdoes FedBE, thanks in part to our proposed \textbf{SynKD} strategy. This approach employs local teacher and student models that learn from local and global representations. Through bidirectional distillation each round, they aim to preserve local inference quality while customizing the student model for local data. The teacher model imparts local insights to the student model, ensuring it isn't overwhelmed by global trends, which could weaken local specificity. The student model, endowed with both local and global insights, also enhances the teacher model, promoting a balanced interchange and safeguarding against representational drift. This strategy improves performance in TFGC tasks.

Additionally, our PRFL improves as more clients participate across datasets. With the exception of the Aircraft-16 dataset, our method beats FedAvg with just 10\% client participation. This success stems from our proposed \textbf{SynKD}, which enables concurrent training and knowledge exchange between the local teacher and student models in one round of updates, enhancing the blend of local and global knowledge. These results demonstrate our method's strong capability for information fusion and its efficient learning and resource use.

\subsection{Ablation Study}
In this subsection, we demonstrate the effectiveness and the necessity of the important components of our method through a series of ablation studies. PRFL's key innovations are in knowledge distillation and parameter decomposition, so we designed two sets of experiments: \textbf{Knowledge Distillation-based Ablation} and \textbf{Parameter Decomposition-based Ablation}. Here's the setup for each:

\paragraph{Knowledge Distillation-based Ablation} We evaluate the impact of our bidirectional knowledge distillation loss by conducting experiments without certain parts:
\begin{itemize}
    \item \textbf{PRFL-A:} This version omits the Distillation Auxiliary Matrix for local models. The correction loss is simply the MSE: $\mathcal{L}_{cor} = \frac{1}{n} \sum_{i=1}^{n}(\mathbf{H}^s_{hs} - \mathbf{H}^t_{hs})$.
    \item \textbf{RPFL-B:} This version removes the Latent Representation Loss. The bidirectional knowledge distillation loss is defined as $\mathcal{L}^t_{bik} = \mathcal{L}^t_{d} + \mathcal{L}^t_{task}$ and $\mathcal{L}^s_{bik} = \mathcal{L}^s_{d} + \mathcal{L}^s_{task}$ for the local teacher and student models, respectively.
\end{itemize}

\paragraph{Parameter Decomposition-based Ablation} We explore the effectiveness of our parameter decomposition strategy:
\begin{itemize}
    \item \textbf{PRFL-C:} This scenario removes Dynamic Parameters Decomposition. As a result, full client model parameters are transmitted between clients and the server instead of just a subset of lightweight local model parameters.
    \item \textbf{PRFL-D:} This version includes Dynamic Parameters Decomposition but lacks the Akaike Information Criterion constraints. The decision on the number of singular values to transmit is based on the following formula:
    \begin{equation}
        \min_K \frac{\sum_{i=1}^K \sigma_i^2}{\sum_{i=1}^Q \sigma_i^2} \textgreater \alpha.
    \end{equation}
\end{itemize}
We conducted ablation studies for four scenarios, consistent with previous experiments, to assess our method at a 10\% client participation rate. These experiments used fine-grained categorical datasets and the results are presented in Tables~\ref{tab:ablation_dir002} (for $\lambda=0.02$), ~\ref{tab:ablation_dir01} (for $\lambda=0.1$), and ~\ref{tab:ablation_path} (for Pathological non-iid).

\begin{table}[H]
  \centering
  \resizebox{.65\textwidth}{!}{
    \begin{tabular}{c|c|c|c|c}
    \toprule
    Method & MTARSI & MSTAR & FGSCR-42 & Aircraft-16 \\
    \midrule
    \multirow{3}[1]{*}{PRFL} & \bf 97.01  & \bf 98.28  & \bf 96.31  & \bf 96.62  \\
       & \small $\pm 0.98$ & \small $\pm 0.78$ & \small $\pm 0.62$ & \small $\pm 1.46$ \\
       & \multicolumn{4}{c}{\cellcolor{red! 20} \bf $8.75\%$ of Complete Model Param.} \\
    \midrule
    \multirow{3}[0]{*}{PRFL-A} & 94.20 & 95.23 & 92.99 & 93.10 \\
       & \small $\pm 0.41$ & \small $\pm 0.54$ & \small $\pm 0.85$ & \small $\pm 0.29$ \\
       & \multicolumn{4}{c}{\cellcolor{red! 20} \bf $8.75\%$ of Complete Model Param.} \\
    \midrule
    \multirow{3}[1]{*}{PRFL-B} & 92.11 & 94.74 & 89.99 & 90.94 \\
       & \small $\pm 0.15$ & \small $\pm 0.34$ & \small $\pm 0.23$ & \small $\pm 0.57$ \\
       & \multicolumn{4}{c}{\cellcolor{red! 20} \bf $8.75\%$ of Complete Model Param.} \\
    \midrule
    \multirow{3}[1]{*}{PRFL-C} & \textcolor{red}{\bf 96.94} & \textcolor{red}{\bf 98.31} & \textcolor{red}{\bf 96.12} & \textcolor{red}{\bf 97.00} \\
       & \small $\pm 0.43$ & \small $\pm 0.42$ & \small $\pm 0.54$ & \small $\pm 0.33$ \\
       & \multicolumn{4}{c}{\cellcolor{green! 20} \bf $100\%$ of Complete Model Param.} \\
    \midrule
    \multirow{3}[1]{*}{PRFL-D} & 95.67 & 96.81 & 95.35 & 94.27 \\
       & \small $\pm 0.79$ & \small $\pm 0.09$ & \small $\pm 0.14$ & \small $\pm 0.25$ \\
       & \multicolumn{4}{c}{\cellcolor{green! 20} \bf $11.26\%$ of Complete Model Param.} \\
    \bottomrule
    \end{tabular}}
    \caption{Comparison of performance under the Dirichlet Non-IID setting ($\lambda= 0.02$), where $\textbf{Bold}$ denotes the best performance of incense under knowledge distillation-based ablation experiments and $\textbf{\textcolor{red}{Bold}}$ denotes the best result under parameters decomposition-based ablation experiments.} 
  \label{tab:ablation_dir002}%
\end{table}

\vspace{-15pt}

\begin{table}[H]
  \centering
  \resizebox{.65\textwidth}{!}{
    \begin{tabular}{c|c|c|c|c}
    \toprule
    Method & MTARSI & MSTAR & FGSCR-42 & Aircraft-16 \\
    \midrule
    \multirow{3}[1]{*}{PRFL} & \bf 90.35  & \bf 97.18  & \bf 84.49  & \bf 87.04  \\
       & \small $\pm 0.79$ & \small $\pm 0.15$ & \small $\pm 0.91$ & \small $\pm 0.87$ \\
       & \multicolumn{4}{c}{\cellcolor{red! 20} \bf $8.75\%$ of Complete Model Param.} \\
    \midrule
    \multirow{3}[0]{*}{PRFL-A} & 87.50 & 93.75 & 82.83 & 81.00 \\
       & \small $\pm 0.42$ & \small $\pm 0.59$ & \small $\pm 0.45$ & \small $\pm 0.79$ \\
       & \multicolumn{4}{c}{\cellcolor{red! 20} \bf $8.75\%$ of Complete Model Param.} \\
    \midrule
    \multirow{3}[1]{*}{PRFL-B} & 86.70 & 91.90 & 81.05 & 79.33 \\
       & \small $\pm 0.30$ & \small $\pm 0.27$ & \small $\pm 0.43$ & \small $\pm 0.58$ \\
       & \multicolumn{4}{c}{\cellcolor{red! 20} \bf $8.75\%$ of Complete Model Param.} \\
    \midrule
    \multirow{3}[1]{*}{PRFL-C} & \textcolor{red}{\bf 90.46} & \textcolor{red}{\bf 97.09} & \textcolor{red}{\bf 84.42} & \textcolor{red}{\bf 87.91} \\
       & \small $\pm 0.66$ & \small $\pm 0.41$ & \small $\pm 0.52$ & \small $\pm 0.49$ \\
       & \multicolumn{4}{c}{\cellcolor{green! 20} \bf $100\%$ of Complete Model Param.} \\
    \midrule
    \multirow{3}[1]{*}{PRFL-D} & 89.27 & 96.16 & 82.98 & 87.01 \\
       & \small $\pm 0.30$ & \small $\pm 0.27$ & \small $\pm 0.43$ & \small $\pm 0.58$ \\
       & \multicolumn{4}{c}{\cellcolor{green! 20} \bf $11.26\%$ of Complete Model Param.} \\
    \bottomrule
    \end{tabular}}
    \caption{Comparison of performance under the Dirichlet Non-IID setting ($\lambda = 0.1$), where $\textbf{Bold}$ denotes the best performance of incense under knowledge distillation-based ablation experiments and $\textbf{\textcolor{red}{Bold}}$ denotes the best result under parameters decomposition-based ablation experiments.} 
  \label{tab:ablation_dir01}%
\end{table}

\vspace{-5pt}

From these tables, we observed that: \textbf{(1)} In the knowledge distillation-based ablation, standard PRFL beats PRFL-A and PRFL-B. This indicates a drop in performance without the Distillation Auxiliary Matrix and Latent Representation Loss, highlighting their importance in our PRFL method; \textbf{(2)} For the parameter decomposition-based ablation, PRFL-C, which sends full model parameters, increased the data exchange by about 91.25\%, leading to more communication overhead and reduced efficiency. The performance gains were marginal or even negative, showing that transmitting an additional 91.25\% of parameters is not cost-effective; (3) PRFL-D also underperformed relative to standard PRFL, with a 2.51\% increase in parameter transmission due to extra redundant information without the AIC constraints. This underscores the value of our singular value selection strategy guided by AIC. In summary, these ablation studies confirm the significance and effectiveness of each component in our PRFL approach.

\vspace{-10pt}

\begin{table}[tbh]
  \centering
  \resizebox{.65\textwidth}{!}{
    \begin{tabular}{c|c|c|c|c}
    \toprule
    Method & MTARSI & MSTAR & FGSCR-42 & Aircraft-16 \\
    \midrule
    \multirow{3}[1]{*}{PRFL} & \bf 89.85  & \bf 96.68  & \bf 80.20  & \bf 84.24  \\
       & \small $\pm 0.37$ & \small $\pm 1.02$ & \small $\pm 0.41$ & \small $\pm 0.29$ \\
       & \multicolumn{4}{c}{\cellcolor{red! 20} \bf $8.75\%$ of Complete Model Param.} \\
    \midrule
    \multirow{3}[0]{*}{PRFL-A} & 86.27 & 94.20 & 77.64 & 83.98 \\
       & \small $\pm 0.42$ & \small $\pm 0.43$ & \small $\pm 0.51$ & \small $\pm 0.04$ \\
       & \multicolumn{4}{c}{\cellcolor{red! 20} \bf $8.75\%$ of Complete Model Param.} \\
    \multirow{3}[1]{*}{PRFL-B} & 85.88 & 93.21 & 77.02 & 82.16 \\
       & \small $\pm 0.17$ & \small $\pm 0.42$ & \small $\pm 0.64$ & \small $\pm 0.08$ \\
       & \multicolumn{4}{c}{\cellcolor{red! 20} \bf $8.75\%$ of Complete Model Param.} \\
    \midrule
    \multirow{3}[1]{*}{PRFL-C} & \textcolor{red}{\bf 90.21} & \textcolor{red}{\bf 96.57} & \textcolor{red}{\bf 79.88} & \textcolor{red}{\bf 85.00} \\
       & \small $\pm 0.98$ & \small $\pm 0.54$ & \small $\pm 0.37$ & \small $\pm 0.42$ \\
       & \multicolumn{4}{c}{\cellcolor{green! 20} \bf $100\%$ of Complete Model Param.} \\
    \midrule
    \multirow{3}[1]{*}{PRFL-D} & 87.31 & 95.02 & 78.88 & 84.12 \\
       & \small $\pm 0.24$ & \small $\pm 0.27$ & \small $\pm 0.13$ & \small $\pm 0.22$ \\
       & \multicolumn{4}{c}{\cellcolor{green! 20} \bf $11.26\%$ of Complete Model Param.} \\
    \bottomrule
    \end{tabular}}
    \caption{Comparison of performance under the Pathological Non-IID setting, where $\textbf{Bold}$ denotes the best performance of incense under knowledge-based distillation experiments and $\textbf{\textcolor{red}{Bold}}$ denotes the best result under parameters decomposition-based ablation experiments.} 
  \label{tab:ablation_path}%
\end{table}

\subsection{Parameters Impact Study}

In this section, we examine the impact of the parameter $\alpha$ on model performance within our parameter decomposition strategy in PRFL. We tested five different values—0.9, 0.92, 0.94, 0.96, and 0.99—in addition to the standard $\alpha=0.98$ used in our experiments. We conducted experiments on four fine-grained classification datasets, with results displayed in Table~\ref{tab:paramstudy}.

\begin{table}[tbh]
  \centering
  \resizebox{.65\textwidth}{!}{
    \begin{tabular}{c|c|c|c|c}
    \toprule
    $\alpha$  & MTARSI & MSTAR & FGSCR-42 & Aircraft-16 \\
    \midrule
    \multirow{2}[1]{*}{0.98} & \bf 90.35  & \bf 97.18  & \bf 84.49  & \bf 87.04  \\
       & \small $\pm 0.72$ & \small $\pm 0.45$ & \small $\pm 0.91$ & \small $\pm 0.82$ \\
    \midrule
    \multirow{2}[0]{*}{0.90} & 89.47  & 96.98  & 83.50  & 86.14 \\
       & \small $\pm 0.82$ & \small $\pm 0.28$ & \small $\pm 0.95$ & \small $\pm 0.74$ \\
    \multirow{2}[0]{*}{0.92} & 89.86  & 97.02  & 83.76  & 86.35 \\
       & \small $\pm 0.46$ & \small $\pm 0.10$ & \small $\pm 0.18$ & \small $\pm 0.60$ \\
    \multirow{2}[0]{*}{0.94} & 89.94  & 97.01  & 83.81  & 86.54 \\
       & \small $\pm 0.38$ & \small $\pm 0.57$ & \small $\pm 0.40$ & \small $\pm 0.92$ \\
    \multirow{2}[0]{*}{0.96} & \underline{90.21}  & 97.11  & 84.14  & 86.70 \\
       & \small $\pm 0.17$ & \small $\pm 0.06$ & \small $\pm 0.59$ & \small $\pm 0.45$ \\
    \multirow{2}[1]{*}{0.99} & 90.11 & \underline{97.12}  & \underline{84.30}  & \underline{86.91} \\
       & \small $\pm 0.95$ & \small $\pm 0.23$ & \small $\pm 0.63$ & \small $\pm 0.46$ \\
    \bottomrule
    \end{tabular}}
    \caption{Experimental results on the performance of PRFL with different $\alpha$, where \textbf{Bold} denotes the optimal result, \underline{Underline} denotes the suboptimal result. $\alpha=0.98$ was the original setting for our previous experiments.}
  \label{tab:paramstudy}%
\end{table}%

The experiments demonstrate that the original $\alpha$ value of 0.98 yields the best classification performance on the datasets. Other values—0.90, 0.92, 0.94, 0.96, and 0.99—lead to a slight decline in performance compared to $\alpha = 0.98$. This decline is attributed to either insufficient or excessive parameters being sent between the client and the server, causing loss of critical information or addition of unnecessary data, respectively, and consequently reducing performance.
\input{table_dp_exp}

\subsection{Differential Privacy}
To safeguard client privacy in our proposed framework, as mentioned in~\citep{chen2023spatial}, we implement differential privacy (DP)~\citep{dwork2006differential} by adding random noise to the gradients during aggregation. We evaluated the performance of our proposed PRFL both with and without DP. The noise, scaled by a factor $\tau$, is added to the shared parameters. We tested $\tau$ values of $\{1e^{-3}, 1e^{-2}, 5e^{-2}\}$ to apply differential privacy at varying levels in our PRFL and compared it with baseline methods.

Table~\ref{tab:dpexp} shows the performance of PRFL and baseline methods on four fine-grained classification datasets under the Dirichlet Non-IID setting (with $\lambda=0.1$), noting that we set the client participation rate at 10\%.

The results in Table~\ref{tab:dpexp} indicate a dip in PRFL's prediction performance after the introduction of DP. Despite this, PRFL still outperforms other baseline methods on all four datasets. Since PRFL only transmits a subset of parameters during client-server communication, adding noise to just these parameters is enough to maintain privacy security. Additionally, the negative impact on performance due to DP is less severe for PRFL compared to baseline methods that exchange full model parameters during client-server communication.

\paragraph{Discussion about Privacy} FL can be at risk of data breaches even though data is not shared directly with other clients during the training of a global model. An attacker might deduce the original data from the gradients sent by a client, particularly when the batch size and local training steps are small. In our PRFL approach, each participant has a local model, consisting of a teacher and a student model for private data training. Our cost-effective parameter decomposition strategy, used during client-server communication, only transmits a subset of parameters. This makes it challenging for attackers to reconstruct the original data from partial gradients.

Moreover, we enhance the strategy with adjustable control coefficients $\alpha$ and AIC-based dynamic constraints. This further complicates potential inference attacks on the raw data, even as the number of training rounds approaches infinity: $e \rightarrow \infty$.

\subsection{Study of Local Updating Steps}
Considering that many remote sensing devices have limited resources, we examined how local update steps affects performance, particularly in resource-constrained settings. Our proposed method offers several benefits: \textbf{(1)} fast convergence; \textbf{(2)} strong generalization ability; \textbf{(3)} strong information fusion capability; \textbf{(4)} efficient learning and resource utilization. 

\begin{table}[tbh]
  \centering
  \resizebox{.9\textwidth}{!}{
    \begin{tabular}{c|ccc|ccc|ccc|ccc}
    \toprule
    \multicolumn{13}{c}{\cellcolor{green! 15} \textsc{Pathological Non-IID Setting}} \\
    \midrule
    \textsc{Dataset} & \multicolumn{3}{c|}{MTARSI} & \multicolumn{3}{c|}{MSTAR} & \multicolumn{3}{c|}{FGSCR-42} & \multicolumn{3}{c}{Aircraft-16} \\
    \midrule
    Method/Local Steps & 10\% & 30\% & 50\% & 10\% & 30\% & 50\% & 10\% & 30\% & 50\% & 10\% & 30\% & 50\% \\
    \midrule
    \multirow{2}[0]{*}{\textsc{FedAvg (Five Steps)}} & 59.60  & 69.55  & 71.65  & 49.60  & 52.55  & 55.40  & 67.20  & 77.15  & 81.25  & 56.10  & 65.50  & \bf 91.30  \\
       & \small $\pm0.42$ & \small $\pm0.56$ & \small $\pm0.29$ & \small $\pm0.41$ & \small $\pm0.59$ & \small $\pm0.42$ & \small $\pm1.11$ & \small $\pm0.09$ & \small $\pm0.27$ & \small $\pm0.63$ & \small $\pm0.24$ & \small $\pm0.49$ \\
    \midrule
        \multirow{2}[0]{*}{\textsc{One Step}} & 85.20 & 88.32 & 89.28 & 90.78 & 92.62 & 91.19 & 75.17 & 79.54 & 80.23 & 78.84 & 82.35 & 83.81  \\
       & \small $\pm0.63$ & \small $\pm0.92$ & \small $\pm1.12$ & \small $\pm0.26$ & \small $\pm0.81$ & \small $\pm0.42$ & \small $\pm0.73$ & \small $\pm1.34$ & \small $\pm1.20$ & \small $\pm0.98$ & \small $\pm1.31$ & \small $\pm0.17$ \\
    \multirow{2}[0]{*}{\textsc{Two Steps}}  & 87.15 & 90.00 & 91.25 & 92.70 & 94.78 & 93.54 & 76.92 & 81.60 & 82.40 & 80.95 & 84.23 & 85.89   \\
       & \small $\pm0.76$ & \small $\pm0.14$ & \small $\pm0.37$ & \small $\pm0.61$ & \small $\pm0.23$ & \small $\pm0.34$ & \small $\pm0.46$ & \small $\pm0.55$ & \small $\pm0.24$ & \small $\pm0.43$ & \small $\pm0.19$ & \small $\pm0.35$ \\
    \multirow{2}[0]{*}{\textsc{Tree Steps}} & \bf 88.96 & \underline{91.96} & \bf 93.38 & \underline{94.70} & \bf 96.79 & \bf 95.44 & \underline{78.16} & \underline{83.37} & \bf 84.18 & \bf 82.60 & \bf 86.07 & \underline{87.67} \\
       & \small $\pm0.89$ & \small $\pm0.42$ & \small $\pm1.24$ & \small $\pm0.06$ & \small $\pm1.10$ & \small $\pm0.33$ & \small $\pm0.95$ & \small $\pm1.43$ & \small $\pm0.18$ & \small $\pm0.77$ & \small $\pm1.11$ & \small $\pm0.92$  \\
    \multirow{2}[0]{*}{\textsc{Four Steps}} & \underline{88.42}  & \bf 92.33  & \underline{93.00}  & \bf 94.97  & \underline{96.23}  & \underline{95.18}  & \bf 78.62  & \bf 83.46  & \underline{83.77}  & \underline{82.54}  & \underline{85.78}  & 87.56 \\
       & \small $\pm 0.95$ & \small $\pm 0.22$ & \small $\pm 0.67$ & \small $\pm 0.09$ & \small $\pm 1.37$ & \small $\pm 0.41$ & \small $\pm 1.08$ & \small $\pm 0.63$ & \small $\pm 0.33$ & \small $\pm 0.74$ & \small $\pm 1.06$ & \small $\pm 0.18$  \\
    \bottomrule
    \end{tabular}}
    \caption{Experimental results of the proposed PRFL under Pathological Non-IID and resource-constrained setting for four fine-grained classification datasets with $\lambda=0.1$ and client participant ratio $\rvr \in \{10\%, 30\%, 50\%\}$, where \textbf{Bold} are the optimal results, and \underline{Underline} represent suboptimal results.}
  \label{tab:path_exp_resource}
\end{table}   

\begin{table}[tbh]
  \centering
  \resizebox{.9\textwidth}{!}{
    \begin{tabular}{c|ccc|ccc|ccc|ccc}
    \toprule
    \multicolumn{13}{c}{\cellcolor{green! 15} \textsc{Dirichlet Non-IID Setting ($\lambda = 0.1$)}} \\
    \midrule
    \textsc{Dataset} & \multicolumn{3}{c|}{MTARSI} & \multicolumn{3}{c|}{MSTAR} & \multicolumn{3}{c|}{FGSCR-42} & \multicolumn{3}{c}{Aircraft-16} \\
    \midrule
    Method/Local Steps & 10\% & 30\% & 50\% & 10\% & 30\% & 50\% & 10\% & 30\% & 50\% & 10\% & 30\% & 50\% \\
    \midrule
    \multirow{2}[0]{*}{\textsc{FedAvg (Five Steps)}} & 57.55  & 67.48  & 69.60  & 47.56  & 50.53  & 53.37  & 65.15  & 76.10  & 79.22  & 54.08  & 63.43  & \bf 90.23  \\
       & \small $\pm1.21$ & \small $\pm1.67$ & \small $\pm1.80$ & \small $\pm0.71$ & \small $\pm1.45$ & \small $\pm2.10$ & \small $\pm2.11$ & \small $\pm1.01$ & \small $\pm0.24$ & \small $\pm1.83$ & \small $\pm0.55$ & \small $\pm3.12$ \\
    \midrule
    \multirow{2}[0]{*}{\textsc{One Step}} &  85.43 & 86.77  &  87.10 &  86.52 &  86.71 & 87.01  &  84.11 &  85.14 & 86.21 &  84.12 & 84.57  &  85.62 \\
       & \small $\pm 0.13$ &  \small $\pm 0.12$ & \small $\pm 0.26$ & $\pm 0.77$ & \small $\pm 0.28$ & \small $\pm 0.33$ & \small $\pm 0.15$ & \small $\pm0.98$ & \small $\pm0.45$ & \small $\pm1.21$ & \small $\pm0.29$ & \small $\pm1.21$ \\
    \multirow{2}[0]{*}{\textsc{Two Steps}} & 86.22  &  86.79 & 87.11  &  86.62 &  86.88 &  87.25  &  84.54 & 85.42  & 86.23  & 84.51  & 84.79  &  85.67 \\
       & \small $\pm0.76$ & \small $\pm0.14$ & \small $\pm0.32$ & \small $\pm0.61$ & \small $\pm0.20$ & \small $\pm0.34$ & \small $\pm0.41$ & \small $\pm0.57$ & \small $\pm0.24$ & \small $\pm0.43$ & \small $\pm0.19$ & \small $\pm0.35$ \\
    \multirow{2}[0]{*}{\textsc{Tree Steps}} & \underline{86.46} & \underline{86.79} &  \bf 87.36 &  \underline{86.75} & \underline{86.90}  & \underline{87.43}  &  \underline{84.62} & \underline{85.78}  &  \bf 87.00 & \underline{84.57} & \underline{85.02}  &  85.99 \\
       & \small $\pm0.25$ & \small $\pm0.87$ & \small $\pm0.19$ & \small $\pm0.64$ & \small $\pm0.31$ & \small $\pm0.32$ & \small $\pm0.37$ & \small $\pm0.51$ & \small $\pm0.44$ & \small $\pm1.26$ & \small $\pm0.47$ & \small $\pm0.11$ \\
    \multirow{2}[0]{*}{\textsc{Four Steps}} & \bf 86.47  & \bf 87.21  &  \underline{87.33} & \bf 87.14 & \bf 87.55  & \bf 87.73  & \bf 84.70 & \bf 86.09  &  \underline{86.82} &  \bf 84.96  &  \bf 85.37 & \underline{86.23} \\
       & \small $\pm0.12$ & \small $\pm0.16$ & \small $\pm0.18$ & \small $\pm0.14$ & \small $\pm0.51$ & \small $\pm0.41$ & \small $\pm0.33$ & \small $\pm0.45$ & \small $\pm0.11$ & \small $\pm0.67$ & \small $\pm0.41$ & \small $\pm0.29$ \\
    \bottomrule
    \end{tabular}}
    \caption{Experimental results of the proposed PRFL under Dirichlet Non-IID and resource-constrained setting for four fine-grained classification datasets with $\lambda=0.1$ and client participant ratio $\rvr \in \{10\%, 30\%, 50\%\}$, where \textbf{Bold} are the optimal results, and \underline{Underline} represent suboptimal results.}
  \label{tab:dir_exp_dot2_resource}
\end{table}       

\begin{table}[tbh]
  \centering
  \resizebox{.9\textwidth}{!}{
    \begin{tabular}{c|ccc|ccc|ccc|ccc}
    \toprule
    \multicolumn{13}{c}{\cellcolor{green! 15} \textsc{Dirichlet Non-IID Setting ($\lambda=0.02$)}} \\
    \midrule
    \textsc{Dataset} & \multicolumn{3}{c|}{MTARSI} & \multicolumn{3}{c|}{MSTAR} & \multicolumn{3}{c|}{FGSCR-42} & \multicolumn{3}{c}{Aircraft-16} \\
    \midrule
    Method/Local Steps & 10\% & 30\% & 50\% & 10\% & 30\% & 50\% & 10\% & 30\% & 50\% & 10\% & 30\% & 50\% \\
    \midrule
    \multirow{2}[0]{*}{\textsc{FedAvg (Five Steps)}} & 30.05  & 37.86  & 39.03  & 37.14  & 48.42  & 48.25  & 41.39  & 46.06  & 50.56  & 38.12  & 42.25  & 42.02  \\
       & \small $\pm 1.13$ &  \small $\pm 0.42$ & \small $\pm 1.26$ & $\pm 2.77$ & \small $\pm 0.78$ & \small $\pm 0.93$ & \small $\pm 0.65$ & \small $\pm1.98$ & \small $\pm1.45$ & \small $\pm2.21$ & \small $\pm1.29$ & \small $\pm2.21$ \\
    \midrule
    \multirow{2}[0]{*}{\textsc{One Step}} &  95.43 & 96.77  &  97.10 &  96.52 &  96.71 & 97.01  &  94.11 &  95.14 & 96.21 &  94.12 & 94.57  &  95.62 \\
       & \small $\pm 0.18$ &  \small $\pm 0.77$ & \small $\pm 0.16$ & $\pm 1.26$ & \small $\pm 0.54$ & \small $\pm 0.03$ & \small $\pm 0.25$ & \small $\pm1.10$ & \small $\pm1.21$ & \small $\pm0.20$ & \small $\pm0.42$ & \small $\pm0.25$ \\
    \multirow{2}[0]{*}{\textsc{Two Steps}} & 96.22  &  96.79 & 97.11  &  96.62 &  96.88 &  97.25  &  94.54 & 95.42  & 96.23  & 94.51  & 94.79  &  95.67 \\
       & \small $\pm1.02$ & \small $\pm2.14$ & \small $\pm1.37$ & \small $\pm1.61$ & \small $\pm1.23$ & \small $\pm0.44$ & \small $\pm0.36$ & \small $\pm0.05$ & \small $\pm0.60$ & \small $\pm0.43$ & \small $\pm0.44$ & \small $\pm0.09$ \\
    \multirow{2}[0]{*}{\textsc{Tree Steps}} & \underline{96.46} & \underline{96.79} &  \underline{97.36} &  \underline{96.75} & \underline{96.90}  & \underline{97.43}  &  \underline{94.99} & \underline{95.78}  &  \bf 97.00 & \underline{94.57} & \underline{95.02}  &  \underline{95.99} \\
       & \small $\pm0.45$ & \small $\pm1.87$ & \small $\pm0.09$ & \small $\pm2.64$ & \small $\pm0.81$ & \small $\pm0.02$ & \small $\pm0.27$ & \small $\pm0.59$ & \small $\pm1.01$ & \small $\pm0.42$ & \small $\pm0.74$ & \small $\pm0.06$ \\
    \multirow{2}[0]{*}{\textsc{Four Steps}} & \bf 96.47  & \bf 97.21  &  \bf 97.33 & \bf 97.14 & \bf 97.55  & \bf 97.73  &  \bf 95.54 & \bf 96.09  &  \underline{96.82} &  \bf 94.96  &  \bf 95.37 & \bf 96.23 \\
       & \small $\pm0.42$ & \small $\pm0.26$ & \small $\pm0.07$ & \small $\pm1.24$ & \small $\pm0.48$ & \small $\pm0.19$ & \small $\pm0.46$ & \small $\pm0.76$ & \small $\pm1.25$ &  \small $\pm1.40$  &  \small $\pm1.00$  & \small $\pm1.11$ \\
    \bottomrule
    \end{tabular}}
    \caption{Experimental results of the proposed PRFL under Dirichlet Non-IID and resource-constrained setting for four fine-grained classification datasets with $\lambda=0.02$ and client participant ratio $\rvr \in \{10\%, 30\%, 50\%\}$, where \textbf{Bold} are the optimal results, and \underline{Underline} represent suboptimal results.}
  \label{tab:dir_exp_dot1_resource}
\end{table}

The result of our method under different settings and local update steps is shown in Table~\ref{tab:path_exp_resource} for Pathological Non-IID, Table~\ref{tab:dir_exp_dot1_resource} for Dirichlet Non-IID with $\lambda=0.1$, and Table~\ref{tab:dir_exp_dot2_resource} for Dirichlet Non-IID with $\lambda=0.02$. Our findings show that our method outperforms FedAvg, even when FedAvg completes five local updates and our method only performs one. This suggests our method converges more quickly and generalizes better.

\section{Conclusion, Limitations and Future Works}
\paragraph{Conclusion} In this paper, we introduced a novel privacy-preserving framework tailored for target fine-grained classification of remote sensing. This framework combines bidirectional knowledge distillation with dynamic parameter decomposition. The bidirectional knowledge distillation allows clients to learn from a rich global knowledge base, enhancing the personalized representation of local data and offering a robust, customized solution for the task in varied environments. Additionally, our dynamic parameter decomposition method is cost-effective, requiring participants to upload only a few parameters rather than the entire model during communication with the server. This is particularly advantageous in resource-constrained settings. Our extensive testing on real-world datasets confirms the effectiveness and advantages of our approach.

\paragraph{Limitations} Our work has two main limitations: \textbf{(1)} Training both the student and teacher models on each client can slightly strain local resources, especially when both models are large; \textbf{(2)} The effect of the low-rank decomposition process-\textit{significantly reducing the size of uploaded transmissions}-may fail when the parameter dimensions of the student models on each client are not widely separated or consistent.

\paragraph{Future Work} In future work, we plan to concentrate on two areas. Firstly, at the framework level, we aim to reduce communication costs between clients and servers while maintaining high performance. Currently, our framework requires training both the student and instructor models on the client side, which can increase local workload. We'll look into using lighter local models within our framework to achieve better performance. Secondly, at the application level, we intend to extend the framework to a wider range of applications, particularly in visual language tasks. This includes, but is not limited to, multimodal remote sensing data retrieval and fusion. Our goal is to offer cost-effective solutions for remote sensing applications on edge devices and other resource-constrained environments.

\bibliography{main}
\bibliographystyle{tmlr}

\end{document}

%% file: math_commands.tex
\usepackage{amsmath,amsfonts,bm}

\def\eqref#1{equation~\ref{#1}}

\def\1{\bm{1}}

\def\rvg{{\mathbf{g}}}

\def\rvp{{\mathbf{p}}}
\def\rvq{{\mathbf{q}}}
\def\rvr{{\mathbf{r}}}

\def\mH{{\bm{H}}}

\def\mW{{\bm{W}}}
\def\mX{{\bm{X}}}
\def\mY{{\bm{Y}}}

\DeclareMathAlphabet{\mathsfit}{\encodingdefault}{\sfdefault}{m}{sl}
\SetMathAlphabet{\mathsfit}{bold}{\encodingdefault}{\sfdefault}{bx}{n}

\def\gL{{\mathcal{L}}}

\def\sR{{\mathbb{R}}}

\def\emU{{U}}
\def\emV{{V}}

\DeclareMathOperator*{\argmin}{arg\,min}

%% file: table_dp_exp.tex
\begin{table}[tb]
  \centering
  \resizebox{.9\textwidth}{!}{
    \begin{tabular}{c|c|c|c|c|c}
    \toprule
    \sc Method & \sc Differential Privacy & MTARSI & MSTAR & FGSCR-42 & Aircraft-16 \\
    \midrule
    \multirow{5}[2]{*}{\sc FedAvg} & \textit{wo}  &  57.55$_{\pm 1.03}$  &  47.56$_{\pm 0.74}$  & 65.15$_{\pm 2.11}$  &  54.08$_{\pm 1.83}$\\
       & \textit{w} ($\tau=1e^{-3}$) &  56.21$_{\pm 0.72}$  &  45.97$_{\pm 0.12}$  &  64.21$_{\pm 1.16}$  & 53.31$_{\pm 0.46}$ \\
       & \textit{w} ($\tau=1e^{-2}$) &  54.64$_{\pm 0.44}$  &  42.30$_{\pm 0.34}$  &  62.77$_{\pm 0.63}$  & 50.56$_{\pm 0.51}$ \\
       & \textit{w} ($\tau=5e^{-2}$) &  47.21$_{\pm 0.21}$  &  38.22$_{\pm 0.93}$  &  52.99$_{\pm 0.46}$  & 43.53$_{\pm 0.78}$ \\
        & \cellcolor{green! 20} Mean Variation &  \cellcolor{green! 20} 8.45 $\downarrow$  &  \cellcolor{green! 20} 11.33 $\downarrow$ & \cellcolor{green! 20} 7.93 $\downarrow$   &  \cellcolor{green! 20} 9.15 $\downarrow$\\
    \midrule
      \multirow{5}[2]{*}{\sc FedBN} &  \textit{wo}  &  73.51$_{\pm 1.04}$  & 66.13$_{\pm 0.06}$  &  81.93$_{\pm 1.12}$  & 61.92$_{\pm 1.81}$ \\
       & \textit{w} ($\tau=1e^{-3}$) &  69.52$_{\pm 0.23}$  &  62.70$_{\pm 0.36}$  &  77.43$_{\pm 0.06}$  & 58.62$_{\pm 1.71}$ \\
       & \textit{w} ($\tau=1e^{-2}$) &  67.98$_{\pm 0.48}$  &  61.07$_{\pm 0.52}$  &  75.75$_{\pm 1.03}$  & 57.25$_{\pm 1.67}$ \\
       & \textit{w} ($\tau=5e^{-2}$) &  65.33$_{\pm 1.80}$  &  58.06$_{\pm 0.02}$  &  73.10$_{\pm 0.45}$  & 55.47$_{\pm 1.61}$ \\
       & \cellcolor{green! 20} Mean Variation &  \cellcolor{green! 20} 4.18 $\downarrow$  &  \cellcolor{green! 20} 4.07 $\downarrow$  &  \cellcolor{green! 20} 4.83 $\downarrow$  & \cellcolor{green! 20} 3.45 $\downarrow$  \\
    \midrule
       \multirow{5}[2]{*}{\sc PerFedAvg} &  \textit{wo}  &  78.30$_{\pm 0.47}$  &  86.31$_{\pm 0.54}$  &  75.95$_{\pm 1.28}$  & 69.01$_{\pm 1.29}$ \\
       & \textit{w} ($\tau=1e^{-3}$) &  76.34$_{\pm 0.55}$  &  84.15$_{\pm 0.73}$  &  74.05$_{\pm 1.92}$  & 67.28$_{\pm 1.26}$ \\
       & \textit{w} ($\tau=1e^{-2}$) &  73.61$_{\pm 0.31}$  &  81.09$_{\pm 0.81}$  &  71.37$_{\pm 1.00}$  & 64.87$_{\pm 0.21}$ \\
       & \textit{w} ($\tau=5e^{-2}$) &  71.62$_{\pm 0.49}$  &  78.60$_{\pm 0.29}$  &  69.51$_{\pm 0.75}$  & 63.11$_{\pm 0.18}$ \\
       & \cellcolor{green! 20} Mean Variation & \cellcolor{green! 20} \underline{3.45} $\downarrow$  &  \cellcolor{green! 20} \underline{3.95} $\downarrow$  &  \cellcolor{green! 20} \underline{3.21} $\downarrow$  &  \cellcolor{green! 20} \underline{2.95} $\downarrow$  \\
    \midrule
      \multirow{5}[2]{*}{\sc FedBE} &  \textit{wo}  &  85.10$_{\pm 0.22}$  & 88.94$_{\pm 0.41}$   &  78.45$_{\pm 1.10}$  & 79.62$_{\pm 0.75}$ \\
       & \textit{w} ($\tau=1e^{-3}$) &  82.97$_{\pm 0.09}$  &  86.71$_{\pm 0.29}$  &  76.49$_{\pm 0.25}$  & 77.62$_{\pm 0.23}$ \\
       & \textit{w} ($\tau=1e^{-2}$) &  80.85$_{\pm 0.26}$  &  84.49$_{\pm 0.38}$  &  74.53$_{\pm 0.60}$  & 75.64$_{\pm 0.51}$ \\
       & \textit{w} ($\tau=5e^{-2}$) &  77.82$_{\pm 0.82}$  &  81.44$_{\pm 0.66}$  &  71.78$_{\pm 1.01}$  & 72.89$_{\pm 0.49}$ \\
       & \cellcolor{green! 20} Mean Variation &  \cellcolor{green! 20} 4.42 $\downarrow$ &  \cellcolor{green! 20} 4.62 $\downarrow$ &  \cellcolor{green! 20} 4.17 $\downarrow$ &  \cellcolor{green! 20} 4.14 $\downarrow$\\
    \midrule
    \multirow{5}[2]{*}{\sc \bf PRFL (Ours)} &  \textit{wo} &  90.35$_{\pm 0.42}$  &  97.18$_{\pm 0.15}$  & 84.49$_{\pm 0.91}$  &  87.04$_{\pm 0.80}$ \\
       & \textit{w} ($\tau=1e^{-3}$) &  89.46$_{\pm 0.48}$  &  96.21$_{\pm 0.99}$  &  83.65$_{\pm 0.40}$  &  86.17$_{\pm 0.56}$ \\
       & \textit{w} ($\tau=1e^{-2}$) &  88.71$_{\pm 0.77}$  &  94.89$_{\pm 0.17}$  &  82.33$_{\pm 0.59}$  &  84.98$_{\pm 0.55}$ \\
       & \textit{w} ($\tau=5e^{-2}$) &  86.33$_{\pm 0.75}$  &  92.73$_{\pm 0.38}$  &  80.27$_{\pm 0.27}$  &  82.87$_{\pm 0.23}$ \\
       & \cellcolor{red! 20} Mean Variation &  \cellcolor{red! 20} \bf 2.01 $\downarrow$ &  \cellcolor{red! 20} \bf 2.15 $\downarrow$ &  \cellcolor{red! 20} \bf 2.22 $\downarrow$ &  \cellcolor{red! 20} \bf 2.17 $\downarrow$\\
    \bottomrule
    \end{tabular}}
    \caption{Results of Differential Privacy experiments (DP factor $\tau \in \{1e^{-3}, 1e^{-2}, 5e^{-2}\}$) of the proposed PRFL and baseline methods under four fine-grained classification datasets under the Dirichlet Non-IID setting ($\lambda=0.1$), where \textbf{Bold} denotes the optimal result, \underline{Underline} denotes the suboptimal result, $\downarrow$ denotes the degradation of the performance, with lower being better. Note that \textit{w} and \textit{wo} present \textit{with} and \textit{without}, respectively.}
  \label{tab:dpexp}
\end{table}%